\documentclass[10pt,twocolumn,letterpaper]{article}

\usepackage[pagenumbers]{wacv} 

\usepackage{graphicx}
\usepackage{amsmath}
\usepackage{amssymb}
\usepackage{booktabs}
\usepackage{xcolor}

\usepackage[pagebackref,breaklinks,colorlinks]{hyperref}

\usepackage[capitalize]{cleveref}
\crefname{section}{Sec.}{Secs.}
\Crefname{section}{Section}{Sections}
\Crefname{table}{Table}{Tables}
\crefname{table}{Tab.}{Tabs.}

\def\wacvPaperID{698} 
\def\confName{WACV}
\def\confYear{2025}

\begin{document}

\title{Beyond Spatial Explanations: Explainable Face Recognition in the Frequency Domain}

\author{Marco Huber$^{1,2}$ and Naser Damer$^{1,2}$\\
$^{1}$ Fraunhofer Institute for Computer Graphics Research IGD, Darmstadt, Germany\\
$^{2}$ Department of Computer Science, TU Darmstadt,
Darmstadt, Germany\\
Email: marco.huber@igd.fraunhofer.de
}
\maketitle

\begin{abstract}
The need for more transparent face recognition (FR), along with other visual-based decision-making systems has recently attracted more attention in research, society, and industry. The reasons why two face images are matched or not matched by a deep learning-based face recognition system are not obvious due to the high number of parameters and the complexity of the models. However, it is important for users, operators, and developers to ensure trust and accountability of the system and to analyze drawbacks such as biased behavior. While many previous works use spatial semantic maps to highlight the regions that have a significant influence on the decision of the face recognition system, frequency components which are also considered by CNNs, are neglected. In this work, we take a step forward and investigate explainable face recognition in the unexplored frequency domain. This makes this work the first to propose explainability of verification-based decisions in the frequency domain, thus explaining the relative influence of the frequency components of each input toward the obtained outcome. To achieve this, we manipulate face images in the spatial frequency domain and investigate the impact on verification outcomes. In extensive quantitative experiments, along with investigating two special scenarios cases, cross-resolution FR and morphing attacks (the latter in supplementary material), we observe the applicability of our proposed frequency-based explanations.
\end{abstract}

\section{Introduction}
\label{sec:intro}
Automatic face recognition (FR) has developed into a technology with a wide range of applications and has become an integral part of many everyday situations. It is used to cross borders at automatic border checkpoints, to unlock personal smartphones, or to verify payments. Among other things, this is due to high accuracy, ease of use, and high convenience for users and operators in many scenarios \cite{DBLP:journals/tcsv/JainRP04}. However, in recent years, drawbacks such as unfair behavior \cite{terhorst2021comprehensive, drozdowski2020demographic, Robinson_2020_CVPR_Workshops} of FR systems and vulnerabilities to different types of attacks \cite{9304898, 9252132, NEURIPS2022_dccbeb7a}, including adversarial attacks, gained attention in the research community. Since modern FR systems highly rely on deep learning with millions of parameters in highly complex models, their outcomes are difficult to understand for humans. Gaining a better understanding of the internal behavior of the models is for various reasons of high interest, especially in biometric systems which are often used in security-critical scenarios and process highly private data \cite{gdpr}. A better understanding could increase trust, highlight unfair behavior, or guide developers to achieve higher-performing and more robust models. In the area of explainable FR, several works have been proposed in recent years \cite{xCos, Huber_2024_WACV, xface, bbox, truebbox, Lu_2024_WACV}.

\begin{figure}[tb]
  \centering
  \includegraphics[height=3.5cm]{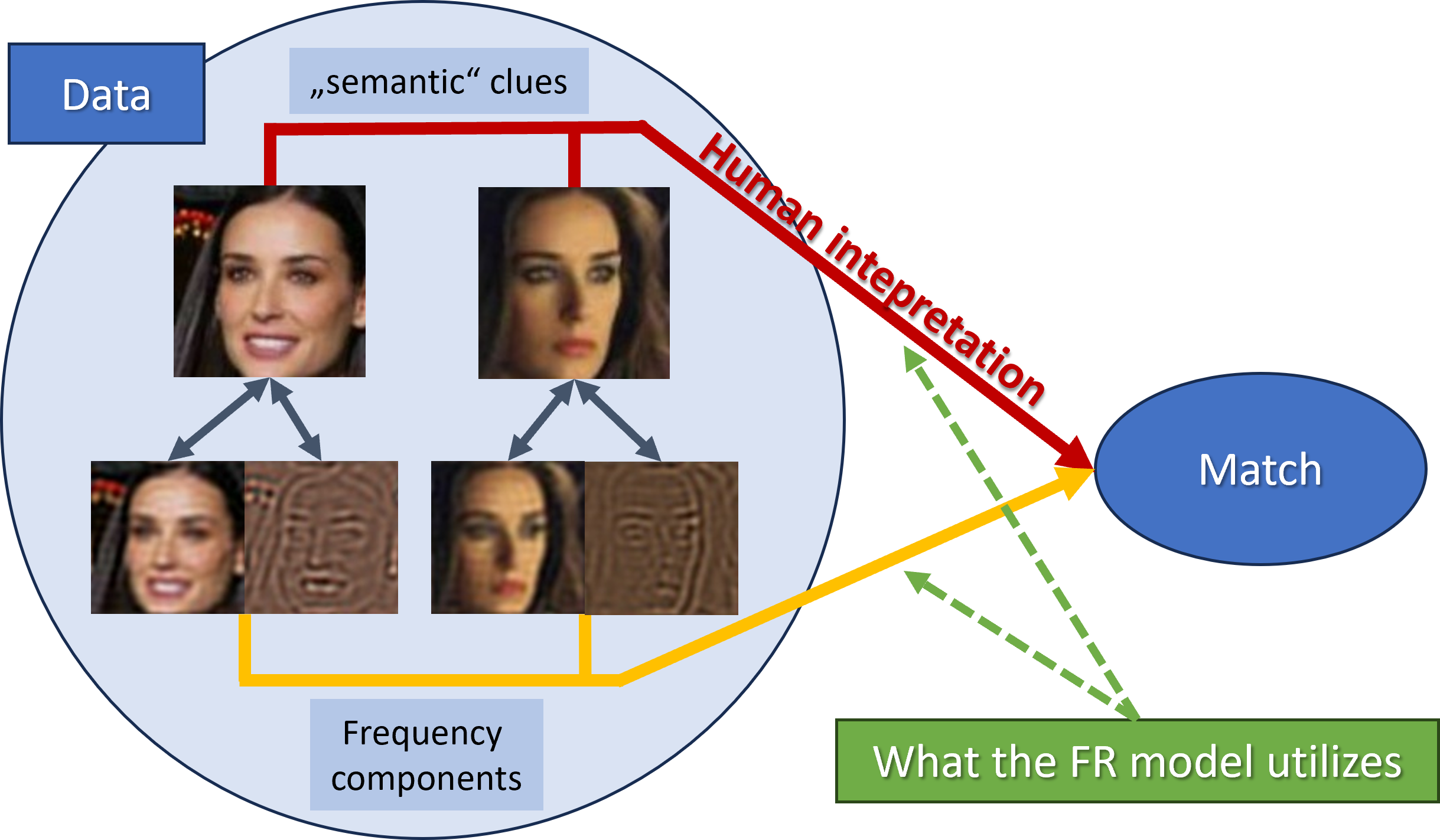}
  \caption{Between the "semantic" components and the frequency components of the data, a correlation exists \cite{wang2020high}. While the human face matcher relies on the human interpretation (semantic component), the FR model also utilizes frequency components \cite{mi2022duetface, mi2023privacy}. Current explainable FR approaches only focus on highlighting the spatial semantic clues.}
  \label{fig:idea}
\end{figure}

Current FR models utilize convolutions neural networks (CNNs) to process face images and distinguish individuals \cite{arcface, elasticface}. These CNNs process the information present in the training face images to create a model that extracts identity representations that can be used to verify and identify individuals by comparing such representations. However, as previous research showed, the CNNs are not only using human-understandable and visible "semantic" clues but also imperceptible high-frequency patterns \cite{wang2020high}. This general idea is visualized in Fig. \ref{fig:idea}. Current approaches \cite{Huber_2024_WACV, xface, bbox} in explainable FR are mainly focused on explanations using spatial saliency maps, which highlight important areas as processed by the FR system. 
However, understanding the influence of different frequency components in FR decisions was never addressed before this work.
For the rest of the paper, for simplicity, we refer to the spatial frequency domain as the frequency domain. 

In the area of biometrics, transparency, explainability, and interpretability of biometric decisions and systems are still considered an outstanding problem \cite{DBLP:journals/tbbis/JainDE22}. In the area of explainable face matching decisions, Lin \etal proposed xCos \cite{xCos}, based on an interpretable cosine metric that provides patched cosine and attention maps, highlighting the important areas for the decision in a spatial manner. More fine-grained explanations are provided by other works \cite{Huber_2024_WACV, xface, bbox, truebbox, Lu_2024_WACV} that provide explanations on pixel or near-pixel level. Recent works aimed at providing textual explanations leveraging large language models \cite{deandres2024good}. In the area of frequencies and their impact on CNNs, Wang \etal \cite{wang2020high} investigated the abilities of CNNs to capture high-frequencies, while Abello \etal \cite{abello2021dissecting} proposed to weigh different frequency bands to gain the influence of different frequency bands for the classification decision. 

In this work, we propose the novel concept of explaining FR in the frequency domain and thus revealing, for any FR verification process, the relative contribution of the frequency components to its outcome.
This allows us to provide more complete explanations, covering also human-imperceptible clues that are used by the FR systems and cannot be visualized in the spatial domain. It is important to highlight, that frequency-based explanations aim at extending the means of explanations and not replace spatial explanations.
To evaluate the performance and show the transferability of our approach, we utilize two different state-of-the-art FR models using adjusted insertion and deletion curves \cite{DBLP:conf/bmvc/PetsiukDS18} and visual investigations.  

Our contributions are: 1) The first work to propose the concept of explaining visual-based decisions of FR in the frequency domain, 2) Investigate frequency-based explainability of FR, thus including human imperceptible clues into the provided explanation, 3) Extensive experiments with two state-of-the-art FR models on the LFW dataset \cite{lfw} in a quantitative and visualized manner, 4) Exploring potential use cases of frequency-based explanations by investigating cross-resolution verifications and morphing attack decisions (in the supplementary material).


\section{Related Work}
The performance of FR systems improved over the recent years and already surpassed the performance of humans \cite{lu2015surpassing}. New FR-specific loss function such as ArcFace \cite{arcface}, CurricularFace \cite{curricularface}, ElasticFace \cite{elasticface} or AdaFace \cite{kim2022adaface}, as well as larger models \cite{resnet} and larger datasets  \cite{DBLP:conf/iccvw/AnZGXZFWQZZF21, DBLP:conf/cvpr/ZhuHDY0CZYLD021} led to this improvement in terms of verification accuracy. A drawback of this is that the increasing size and complexity of the FR models result in a lower understanding of their decisions. This raises questions about the inner workings and causes for decisions, thus motivating innovations targeting explainable FR.

The major research direction in the area of explainable computer vision is the trend of visualizing important areas in images using saliency or heatmaps \cite{selvaraju2017grad, DBLP:conf/cvpr/WangWDYZDMH20}. They are mainly designed and applied to explain classification decisions by highlighting which areas led to a certain classification output. Applying methods like GradCam \cite{selvaraju2017grad} or Score-CAM \cite{DBLP:conf/cvpr/WangWDYZDMH20} on FR models is not natively possible, since they are only designed for classification systems and not for an FR process that includes feature extraction and feature comparison sub-steps. 

This motivated the development of approaches specifically designed to visualize important areas for an FR model's decision. Lin \etal \cite{xCos} proposed a learnable module named xCos which is based on a novel similarity metric and provides patched attention and similarity maps. More fine-grained explanations are generated by the black-box approaches \cite{bbox, truebbox, xface, Lu_2024_WACV}. All of them are based on perturbations and manipulation of the input image in the spatial domain while observing the output/decision. In contrast, to black-box attacks which assume no access to the model weights, also white-box approaches exist. In that direction, Huber \etal \cite{Huber_2024_WACV} proposed to use similarity score-based gradients to obtain fine-grained similarity and dissimilarity heatmaps, while Xu \etal \cite{DBLP:journals/corr/abs-2403-04549} proposed a white-box approach based on feature-guided gradient backpropagation. All existing works explaining FR decisions or other computer vision decisions have focused on the spatial domain and have not addressed explanations in the frequency domain, i.e. answering questions regarding the most important frequency bands contributing to a certain decision. This is of special interest, as the relationship between the frequency spectrum of images and the behavior of CNNs has been recently investigated by Wang \etal \cite{wang2020high}. They especially look at the generalization behavior of CNNs and the impact of the low- and high-frequency components during training and inference.  One of their conclusion is that CNNs capture high-frequency patterns that misalign with human visual preference. 

\begin{figure*}[h]
  \centering
  \includegraphics[width=0.7\textwidth]{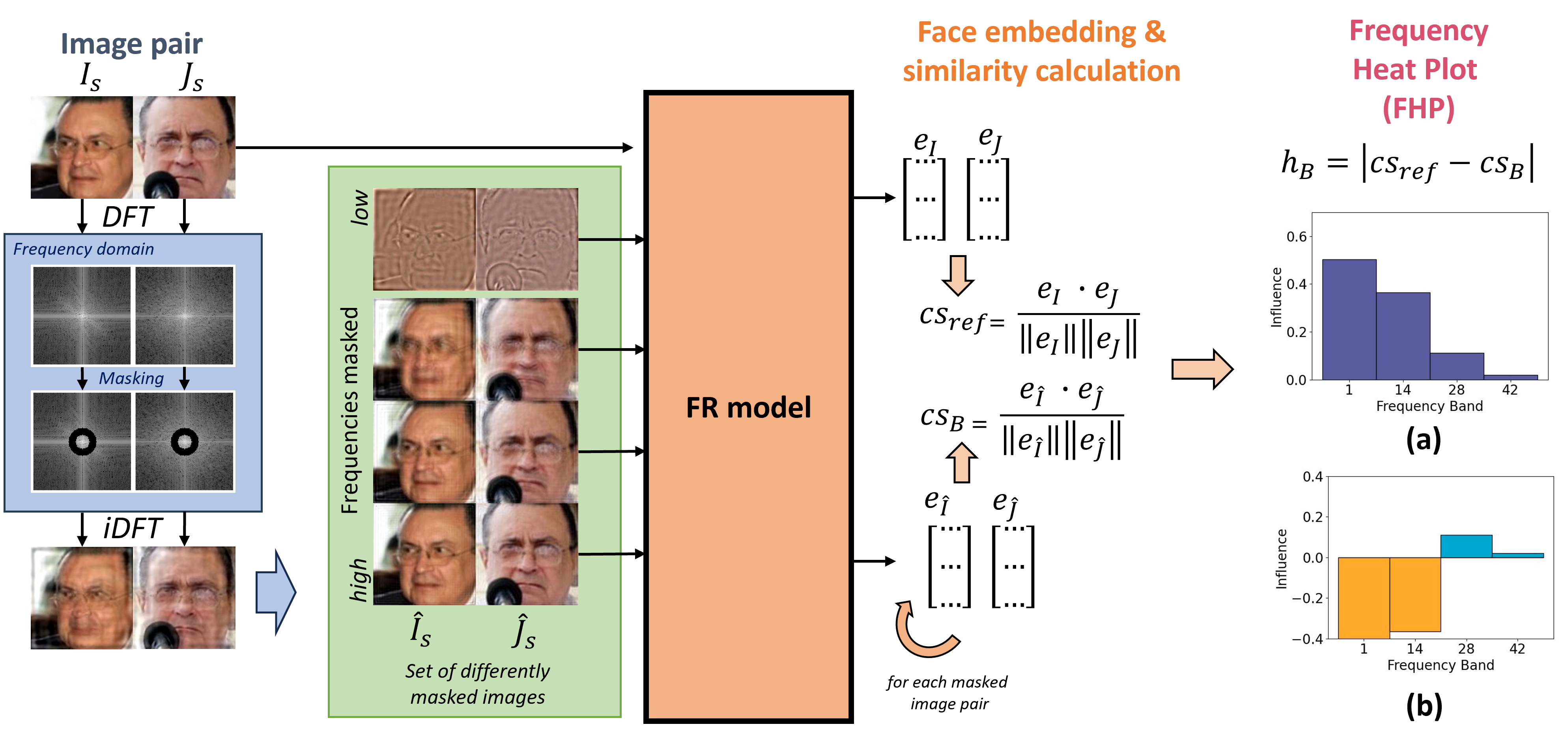}
  \caption{Overview of the proposed frequency-based explainability approach. In the first step, the images of an image pair are transformed into the frequency domain. Certain frequencies are masked and the images are re-transformed into the spatial domain, in a lossless process. In the next step, the unaltered images and the set of frequency-masked images are processed by an FR model to create face embeddings and to calculate cosine similarity scores. In the last step, the difference between the cosine similarity scores of the masked image pairs and the unaltered image pair is used to assign an influence score to the different frequencies (bands). The normalized influences are either presented as the absolute (a) or directed (b) frequency heat plots (FHPs).}
  \label{fig:overview}
\end{figure*}

While not directly referring to frequencies, Ilyas \etal \cite{ilyas2019adversarial} investigated and provided a theoretical framework regarding the existence of non-robust features which are derived by the model from the data distribution and are highly predictive for the model, but incomprehensible to humans. They concluded, that as long as models rely on non-robust, human incomprehensible features the explanations of model behavior cannot be both faithful and human-meaningful \cite{ilyas2019adversarial}. Abello \etal \cite{abello2021dissecting} provided a framework for dividing the frequency spectrum in equally sized disjoint energy discs to investigate the importance of the different discs. Their experiments and the results indicate no clear bias to high-frequencies. Abello \etal \cite{abello2021dissecting} utilized an FR model in their experiments in an identity classification manner, but neglected the established feature creation and matching process. More recently, several works \cite{mi2023privacy, wang2022privacy, mi2022duetface, ji2022privacy} exploited the frequency domain to perform privacy-preserving FR by manipulating the frequencies in different ways. Which points out the importance of frequency domain analyses in FR. However, no works have addressed the explainability task in the frequency domain.

\section{Frequency-based Explanations}
In this section, we propose our frequency-based explainability approach. Utilizing recent findings on the impact and importance of frequencies in the internal behavior of CNNs \cite{wang2020high, abello2021dissecting}, we move away from spatial explanations such as heatmaps and saliency maps \cite{xface, Huber_2024_WACV, truebbox} and explain face matching in the frequency domain. 

The benefit of leveraging the frequency domain for explanation is that it also allows us to understand the relative contributions of different frequency components, including high-frequency patterns used by the model in the explanations that are human imperceptible \cite{wang2020high}. 
To achieve this, we propose to analyze spatial image pairs in the frequency domain using Discrete Fourier Transform (DFT) and apply a black-box masking approach by masking different frequency bands to remove the information present in these frequencies. After the masking, the manipulated image is transformed back to the spatial domain utilizing the inverse Discrete Fourier Transform (iDFT) without any additional information loss. To obtain our explanation in the form of a frequency heat plot (FHP), the change in the comparison score is observed between the frequency-masked image pairs and the unaltered image pair. If the change in the similarity score is relatively large, the masked frequencies had a relatively high influence on the face verification process and vice versa. A schematic overview of our proposed approach is shown in Fig. \ref{fig:overview}.
Our proposed training-free approach follows the black-box approach \cite{xface, truebbox, bbox} which does not need access to the model's internals and can be applied to any CNN-based FR model. 

\subsection{From Spatial Domain to Frequency Domain and Back}
\
To investigate the impact of certain frequencies on the model's behavior, we propose to remove these frequencies from the image pair. Therefore, in the first step, we transform the image from the spatial domain into the frequency domain. Given an image in spatial domain $I_s \in R^{N \times N }$, we apply all following operations color channel-wise, following \cite{abello2021dissecting, wang2020high}. With $I_{s} (x,y)$ we refer to the value in the spatial domain at position $(x, y)$. We refer to the Fourier transform $\mathcal{F}$ as the Discrete Fourier Transform (DFT): 

\small
\begin{equation}
    \mathcal{F}(k, l) = \frac{1}{N^2}\sum^{N-1}_{x=0} \sum^{N-1}_{y=0} I_{s} (x,y) e^{-i2\pi(\frac{kx}{N} + \frac{ly}{N})},
\end{equation}

where $\mathcal{F}(k, l)$ denotes the value of $I_{s} (x,y)$ in the spatial frequency domain. Therefore, $\mathcal{F}(k, l)$ is a complex-valued matrix, where each $(k,l)$ pair represents the frequency. To simplify the calculation of frequency bands later on, we shift the zeroth frequency to the center as it is usually done when processing images in frequency domain, leaving any information unchanged \cite{abello2021dissecting}.

To estimate the influence of each frequency band, we apply a masking scheme. This masking scheme removes all information in a certain frequency band by setting the corresponding value to zero in the frequency domain. Due to the frequency shifting of the zeroth frequency to the center, the size/height  ("low" and "high") of the frequencies of the image in the frequency domain is a norm of $\mathcal{F}(k, l)$. To mask different frequencies, we define a frequency mask $M_{\mathcal{B}} \in R^{N \times N}$ for a frequency band $\mathcal{B}(b,t)$ with $b$ denoting the lower frequency bound and $t$ the upper frequency bound as:

\small
\begin{equation}
    M_\mathcal{B}(k,l) = 
    \begin{cases}
    0, & \text{if $\lVert \mathcal{F}(k, l) \rVert > b$ or $\lVert \mathcal{F}(k, l) \rVert \leq t$}\\ 
    1, & \text{otherwise} 
    \end{cases}, 
\end{equation}

Additionally, we define $s=t-b$ as the frequency band size. We apply the obtained map $M_\mathcal{B}(k,l)$ channel-wise (per color channel) on the image in the frequency domain $\mathcal{F}(k,l)$: 
\small
\begin{equation}
    \mathcal{F}_{M}(k, l) = \mathcal{F}(k, l) \times M_{\mathcal{B}}(k,l),
\end{equation}
to remove the selected frequency band $\mathcal{B}(b,t)$ from the frequency domain image $\mathcal{F}(k, l)$, obtaining the masked version $\mathcal{F}_{M}(k, l)$. In the last step, we invert the shifting of the zeroth frequency to the center and utilize the inverse Discrete Fourier Transform (iDFT), $\mathcal{F}^{-1}$:

\small
\begin{equation}
    I_{s}(x,y) = \mathcal{F}^{-1} = \sum^{N-1}_{k=0} \sum^{N-1}_{l=0}  \mathcal{F}(k, l) e^{i2\pi(\frac{kx}{N} + \frac{ly}{N})},
\end{equation}

to transform the frequency-masked image back to the spatial domain. To summarise briefly, we calculate:
\small
\begin{equation}
    \hat{I}_{s} = \mathcal{F}^{-1}(\mathcal{F}(I_s) \times M_{B}),
\end{equation}
to obtain a spatial version of the image $I_s$, $\hat{I_{s}}$ that had certain frequencies masked ($\mathcal{B}(b,t)$) in the frequency domain.

By doing the above mentioned transformation and masking in the frequency domain for each image on each frequency band $\mathcal{B}_{i}$, we obtain a set of images $\mathcal{I}_{\hat{I}_s}$ of image $I_s$ with certain frequencies, and the information present in this frequencies, removed. It is important to note, that we do not mask the zeroth frequency (DC component) as it carries a different meaning with the average intensity of the pixels in the image and that the transformation step does not change the image without masking.

\subsection{Assigning Frequency Influence for Face Verification}
Starting from the set of images $\mathcal{I}_{\hat{I}_s}$ obtained using the non-overlapping frequency masking, we investigate the change in the similarity of the masked image pairs and the unaltered image pairs given some arbitrary FR model as a measure of influence for each frequency band. 

Therefore, given $\mathcal{I}_{\hat{I}_s}$  and an FR model, $E$, that maps its spatial input $I$ into a feature representation $e_{I}$ in an identity embedding space:

\small
\begin{equation}
    e_{I} = E(I), 
\end{equation}
where $e_I$ is a mathematical representation of the identity present in image $I$ in an identity space learned by the FR model $E$, we calculate for each manipulated image pair $\hat{I_s} \in \mathcal{I}_{\hat{I}_s}, \hat{J}_s \in \mathcal{J}_{\hat{J}_s}$ their corresponding $e_{\hat{I}}$ and $e_{\hat{j}}$.  To calculate the similarity of the representations, we use cosine similarity $cs$, which can be defined as:

\small
\begin{equation}
    cs(e_{I}, e_{J}) = \frac{e_{I} \cdot e_{J}}{\lVert e_{I} \rVert \lVert e_{J} \rVert },
\end{equation}

and provides a comparison score $cs \in [-1,1]$ with values close to 1 indicating similarity, values close to 0 independence, and values close to -1 indicating dissimilarity. The similarity is always calculated between the manipulated image with the same frequency mask applied ($M_{B}$) and we manipulate both images of the pair and not just one, to remove the information present in the frequency from both images and not just from one.
To obtain a reference similarity that we can use to measure the difference in similarity, we calculate the cosine similarity $cs_{ref}$ of the unaltered image pair $I_s, J_s$. In the final step, we define the frequency band influence as:

\small
\begin{equation}
    h_{B} = |cs_{ref} - cs_{B}|
\end{equation}
where $h_{B}$ denotes the influence of a specific frequency band $B$ and $cs_B$ the comparison score of the embedded images $\hat{I_s}$ and $\hat{J_s}$ that had the frequency band $B$ masked. Therefore, we interpret the absolute distance in the cosine similarity score between the unaltered image pair and their $B$-frequency manipulated counterpart as the influence.

\subsection{Visualizing Frequency Influence}

\label{sec:vis}
To visualize the obtained frequency influence scores, we propose two different frequency heat plots (FHPs), the absolute FHP and the directed FHP. The proposed plots are bar plots that visualize the impact of each frequency band on the reference comparison score of the unaltered reference images in terms of similarity. This allows us to visualize the influence of certain frequency bands on the utilized FR models and allows us to also cover high-frequencies in the explanations. To make them comparable between different FHPs we normalize the influence score to sum up to 1 in each plot. While the absolute FHP is based on the absolute distance $H_{b}$, the directed FHP maintains the direction (sign) of the difference after being normalized. This allows us to also observe and explain if a certain frequency band had a positive or negative influence on the similarity. An example for both, the absolute (a) and the directed FIP (b) is shown in Fig. \ref{fig:overview}.

\section{Experimental Setup}
\subsection{Face Recognition Models \& Evaluation Datasets}

To demonstrate the generalizability and validity of our proposed frequency-based explainable FR approach, we utilize two different state-of-the-art FR models. All models share the same ResNet-100 \cite{resnet} architecture and have been trained with the corresponding loss functions. The utilized models are ElasticFace-Arc \cite{elasticface} and CurricularFace \cite{curricularface}. We obtained the trained models (trained on MS1MV2 \cite{arcface}) from the respective official repositories. 

For the evaluation benchmark, following FR explainability works \cite{xface, Huber_2024_WACV}, we utilize Labeled Faces in the Wild (LFW) \cite{lfw} which is an established face verification benchmark and consists of 3000 genuine (matching) and 3000 imposter (non-matching) pairs. The images have been aligned and cropped using MTCNN \cite{DBLP:journals/spl/ZhangZLQ16} following Deng \etal \cite{arcface} to the size of 112 $\times$ 112 $\times$ 3. 

Last, we investigate two possible use cases of our frequency-based explanations: cross-resolution FR and morphing attack decisions (in the supplementary material). For the analysis of the cross-resolution FR, the low-resolution images have been created by down- and up-scaling \cite{DBLP:journals/tifs/LiPMF19} the image with bilinear interpolation using the downscaling factor $m = 0.25$. Our approach should be able to explain, how in low-resolution image pairs less details affect the influence of information present in certain frequencies. We expect that the influence of low frequencies increases as there is less information in high-frequency bands due to the removed details. The details of the morphing attack decisions are provided in the supplementary material.

\subsection{Hyper-parameters}

Based on the approach described above, we investigate different hyper-parameters in our experiments. Motivated by related works, we investigate $\mathcal{L}_1$ (used by Abello \etal \cite{abello2021dissecting}) and $\mathcal{L}_2$ (used by Wang \etal \cite{wang2020high}) norm as the distance functions during the frequency masking. The results using $\mathcal{L}_2$ are provided in this paper, and the results using $\mathcal{L}_1$ are provided in the supplementary material. Furthermore, we chose the size of the frequency band $s \in [1,2,4,8,14]$ to investigate a wide range of possible bandwidths. With $s=1$ we investigate single frequencies, while larger $s$ might provide more robust explanations as single frequencies might have an insignificant effect \cite{abello2021dissecting}.

\subsection{Insertion \& Deletion Curves}

To determine the quality of the produced frequency-based explanations, we utilize adapted insertion and deletion curves \cite{DBLP:conf/bmvc/PetsiukDS18}. The rationale of insertion and deletion curves is, that if the most important areas of an image are removed from an image or inserted into a blank canvas, the performance should decrease or increase most, in contrast to if the same is done for less important areas. This approach has been established as a way to evaluate spatial explainable FR approaches \cite{Lu_2024_WACV, Huber_2024_WACV, huber2022evaluating} along with explainability approaches in other domains \cite{DBLP:conf/bmvc/PetsiukDS18, DBLP:conf/cvpr/WangWDYZDMH20}. We adapt insertion and deletion curves and in contrast to existing works, perform the masking in the frequency domain. Our adapted evaluation aims at estimating the influence of the different frequency bands and then iteratively masking the different frequency bands based on the estimated influence and evaluating using the face verification performance. We perform the masking in the frequency domain on both images of each pair to remove the information present in the frequency from both images. In the evaluation using the insertion and deletion curves, we utilize the influence provided by the absolute FHPs as in contrast to the directed FHPs, they provide a fixed ordering, from highest to lowest influence.

We provide two evaluation scenarios for each curve. In the first case, we investigate the Equal Error Rate (EER) \cite{iso_metric}, which provides the evaluation at the threshold that achieves the same False Match Rate (FMR) as the False Non-Match Rate (FNMR). With the EER we can measure both errors within a single evaluation curve. In the second case, we imitate a more practical scenario with a fixed threshold, that is not adapted during the evaluation. Employed FR systems usually utilize a fixed threshold that is set based on some reference dataset and then used during operation. We set this threshold at FMR$=0.1$ on the unaltered LFW \cite{lfw} dataset and report the error in terms of FNMR.

A fair comparison of frequency-based explanations with spatial explanations using insertion \& delete curves is not possible, as the amount of information removed or inserted is different when manipulating frequencies or pixels. To validate our results, we compare our explanations with a baseline of unaware explanations. These unaware explanations are built by assuming that the frequency bands are masked randomly but using the same frequency band size $s$ as in the proposed approach. The masked frequency bands are in random order for each image pair. In the resulting insertion curves, a fast increase in verification performance indicates a good explanation as it indicates that the approach has captured the most crucial information, and adding these parts should reduce the verification error more than adding less crucial information. In the deletion curves, a fast decrease in verification performance indicates a good explanation as by removing the most crucial information the model should be disturbed more than by removing less crucial information.

\begin{figure*}
     \centering
     \begin{subfigure}[b]{0.24\textwidth}
         \centering
         \includegraphics[width=\textwidth]{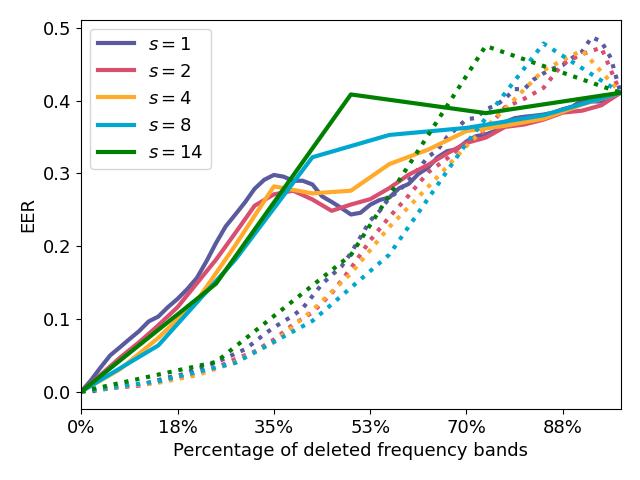}
         \caption{Deletion - EER}
         \label{fig:elaDelEER}
     \end{subfigure}
     \begin{subfigure}[b]{0.24\textwidth}
         \centering
         \includegraphics[width=\textwidth]{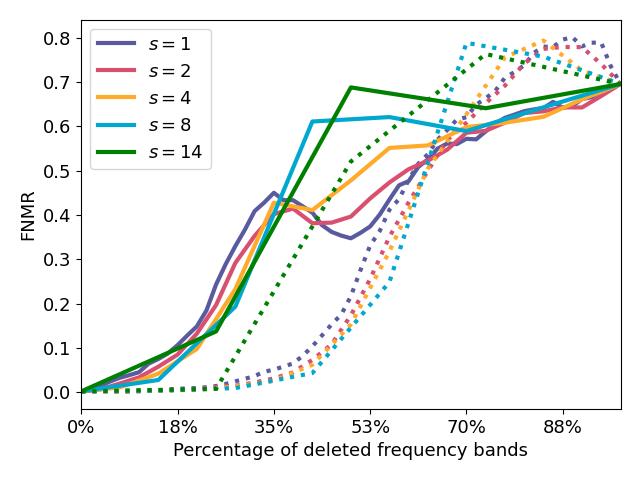}
         \caption{Deletion - FNMR}
         \label{fig:elaDelFNMR}
     \end{subfigure}
     \begin{subfigure}[b]{0.24\textwidth}
         \centering
         \includegraphics[width=\textwidth]{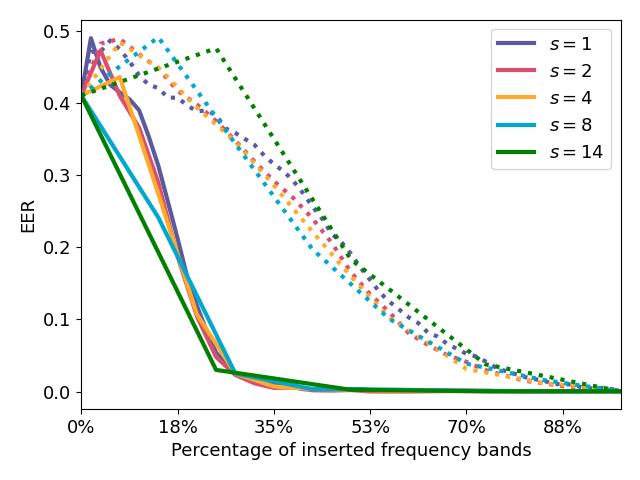}
         \caption{Insertion - EER}
         \label{fig:ealInsEER}
     \end{subfigure}
     \begin{subfigure}[b]{0.24\textwidth}
         \centering
         \includegraphics[width=\textwidth]{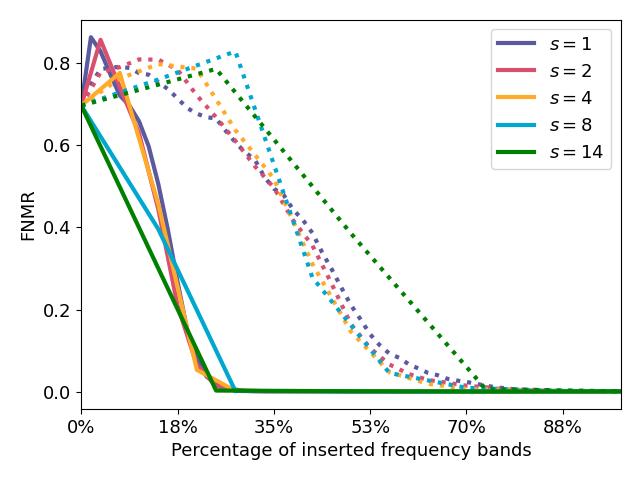}
         \caption{Insertion - FNMR}
         \label{fig:elaInsFNMR}
     \end{subfigure}
        \caption{Deletion and insertion curves using ElasticFace-Arc \cite{elasticface} on LFW \cite{lfw}. The solid lines are a result of our proposed explanations. The dotted line indicates the performance of the baseline with the same frequency band size $s$ as its color counterpart in solid line. Both faster ascending deletion curves and faster descending insertion curves point to the effectiveness of the proposed explanations. }
        \label{fig:quantElastic}
\end{figure*}

\begin{figure*}
     \centering
     \begin{subfigure}[b]{0.24\textwidth}
         \centering
         \includegraphics[width=\textwidth]{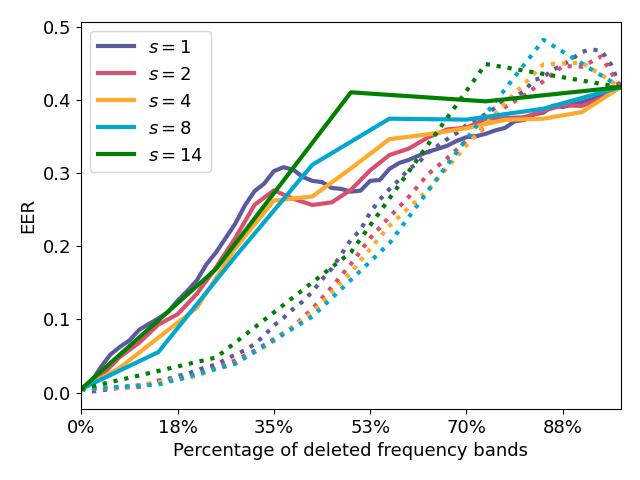}
         \caption{Deletion - EER}
         \label{fig:curDelEER}
     \end{subfigure}
     \begin{subfigure}[b]{0.24\textwidth}
         \centering
         \includegraphics[width=\textwidth]{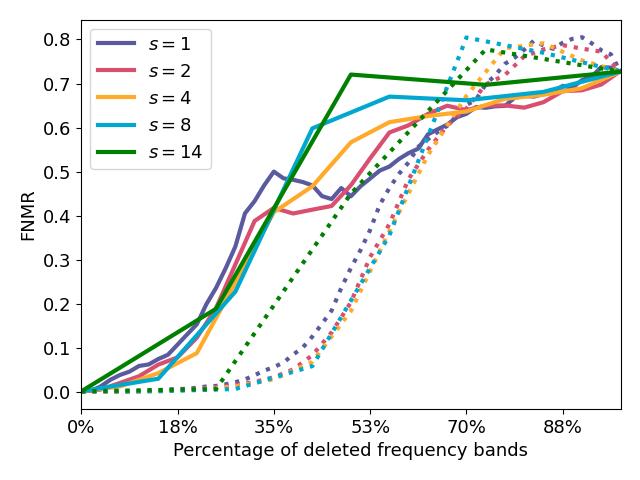}
         \caption{Deletion - FNMR}
         \label{fig:curDelFNMR}
     \end{subfigure}
     \begin{subfigure}[b]{0.24\textwidth}
         \centering
         \includegraphics[width=\textwidth]{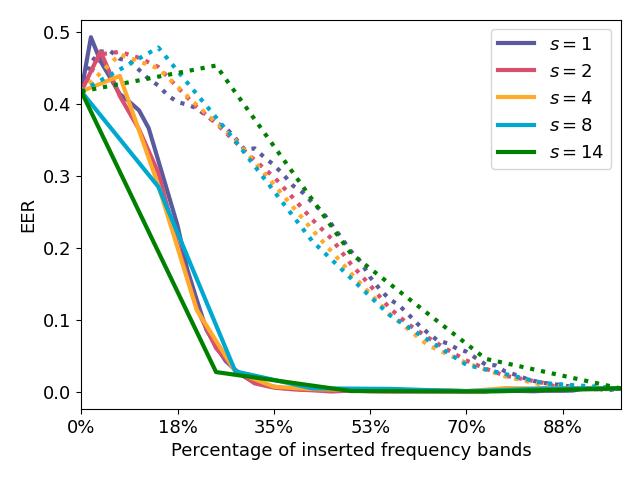}
         \caption{Insertion - EER}
         \label{fig:curInsEER}
     \end{subfigure}
     \begin{subfigure}[b]{0.24\textwidth}
         \centering
         \includegraphics[width=\textwidth]{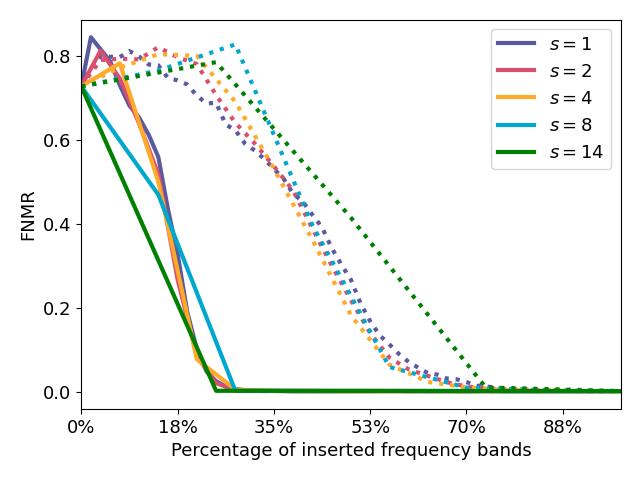}
         \caption{Insertion - FNMR}
         \label{fig:curInsFNMR}
     \end{subfigure}

        \caption{Deletion and insertion curves using CurricularFace \cite{curricularface} on LFW \cite{lfw}. The solid lines are a result of our proposed explanations. The dotted line indicates the performance of the baseline with the same frequency band size $s$ as its color counterpart in solid line. Both faster ascending deletion curves and faster descending insertion curves point to the effectiveness of the proposed explanations.}
        \label{fig:quantCurr}

\end{figure*}

\subsection{Visual Investigations}

For the visual analysis, we present the absolute and directed frequency heat plots (FHP) for image pairs. The FHPs provide an explanation, of how influential certain frequency bands are for the FR model. To demonstrate the usefulness of our proposed FHPs, we provide them for two different use cases. For the cross-resolution use case, we provide absolute and directed FHPs, for the morphing attack decision use case, we provide the mean FHPs and the standard deviation over the genuine and attack pairs in the supplementary material.

\section{Results}
This section presents the results of our proposed frequency-based explainability framework, starting with the presentation of the quantitative findings, followed by the analysis of selected visual observations. In our quantitative analysis, we examine the insertion and deletion curves pertaining to our frequency-based explainable FR approach, as described earlier, employing two distinct FR models and two evaluation scenarios. These scenarios include one with a pre-determined starting threshold (fixed FMR) and another with a dynamically adaptable threshold set by the EER. For the visual investigation, we observe the FHPs of unaltered image pairs as defined by the LFW benchmark \cite{lfw} and low-resolution versions of these pairs. The quantitative results are based on using $L2$-norm as motivated by Wang \etal \cite{wang2020high}, the results using $L1$-norm as motivated in \cite{abello2021dissecting}, are provided in the supplementary material.

\subsection{Quantitative Results}

\begin{figure*}[hbt!]
  \centering
  \includegraphics[width=0.8\textwidth]{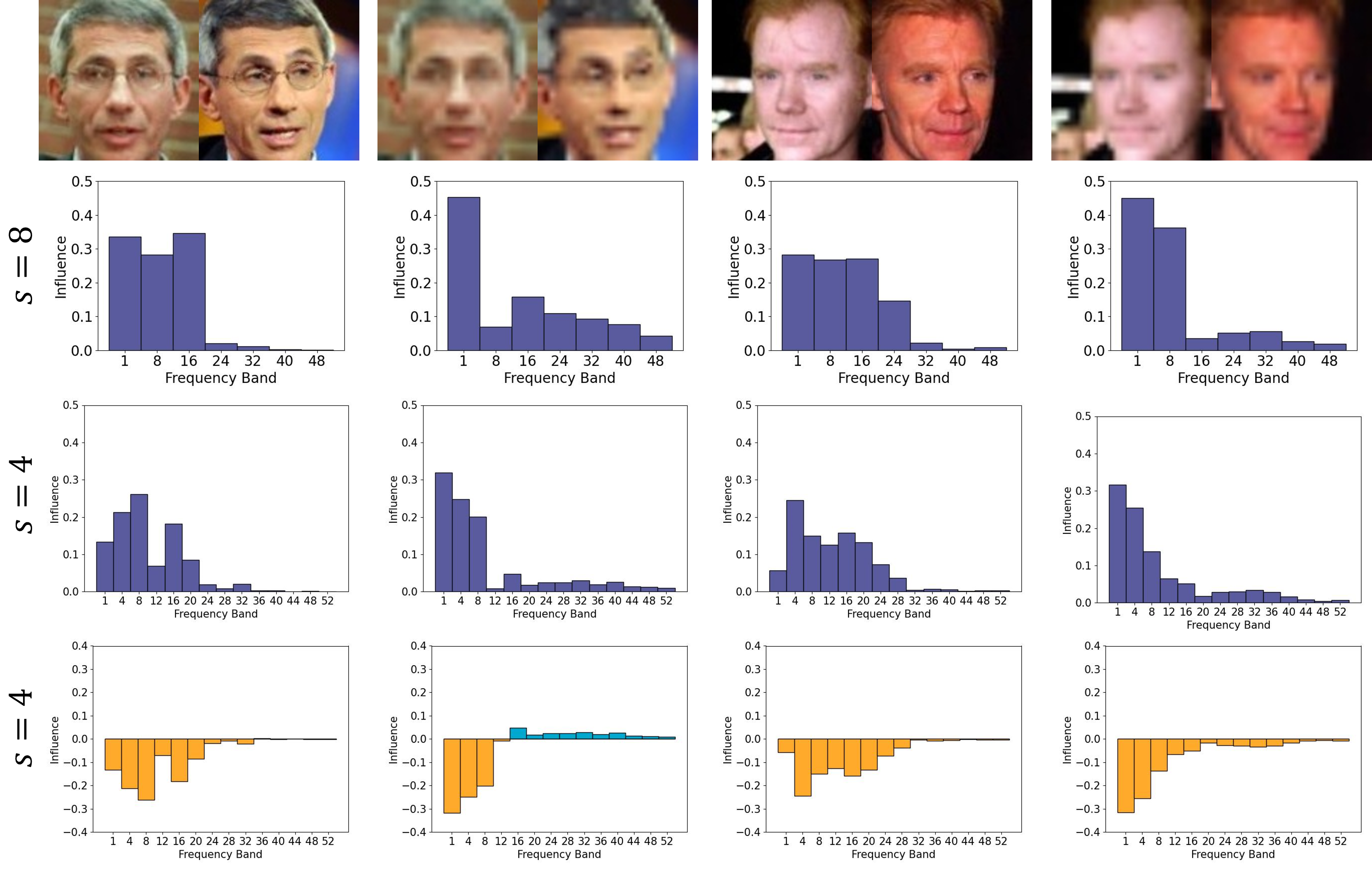}
  \vspace{-2mm}
  \caption{Matching (genuine) pairs and their low-resolution versions with FHPs. The FHPs provide image pair specific frequency-based explanations on the influence of certain frequency bands. The low-frequency bands are the most influential and get even more influential on low-resolution images that lack details. Under each image pair we show two) with different $s$ absolute FHPs followed by a third directed FHP.}
  \label{fig:qual1}
\end{figure*} 

The insertion and deletion curves on the ElasticFace-Arc \cite{elasticface} model for the EER and FNMR case are shown in Fig. \ref{fig:quantElastic} using $L2$-norm. The dotted lines indicate the performance of the baseline. The plot in Fig. \ref{fig:elaDelEER} proves that the explanations by our proposed approach are meaningful, as the error in terms of EER increases faster as the baseline. The frequencies that are determined to have a higher influence have a higher impact on the performance than if they are selected randomly. This remains true for all the considered frequency band sizes and is also true in the insertion curve (Fig. \ref{fig:curInsEER}), where the error decreases faster as our proposed approach successfully detected the most influential frequencies first which reduces the verification error faster than the baseline. 
Similar behavior can also be observed in the FNMR evaluation scenario using a fixed starting threshold based on the FMR (Fig. \ref{fig:elaDelFNMR} and Fig. \ref{fig:elaInsFNMR}).

The insertion and deletion curves on the CurricularFace \cite{curricularface} model are shown in Fig. \ref{fig:quantCurr}. The performance observed in all cases, EER (Fig. \ref{fig:curDelEER}, Fig. \ref{fig:curInsEER}) and fixed starting threshold (Fig. \ref{fig:curDelFNMR}, Fig. \ref{fig:curInsFNMR}) outperforms the baseline and follows a similar performance that can be observed in the ElastiFace-Arc \cite{elasticface} model, which shows the generalizability of the proposed explainability solution. 

Interestingly, in both figures, it can be observed, that even large frequency bands ($s=14$) allow a good estimation of which frequency bands are important, while smaller frequency band sizes allow a more fine-granular explanation of the influence.

\subsection{Visual Investigation - Use Case: Cross Resolution FR}

\begin{figure*}[h]
  \centering
  \includegraphics[width=0.8\textwidth]{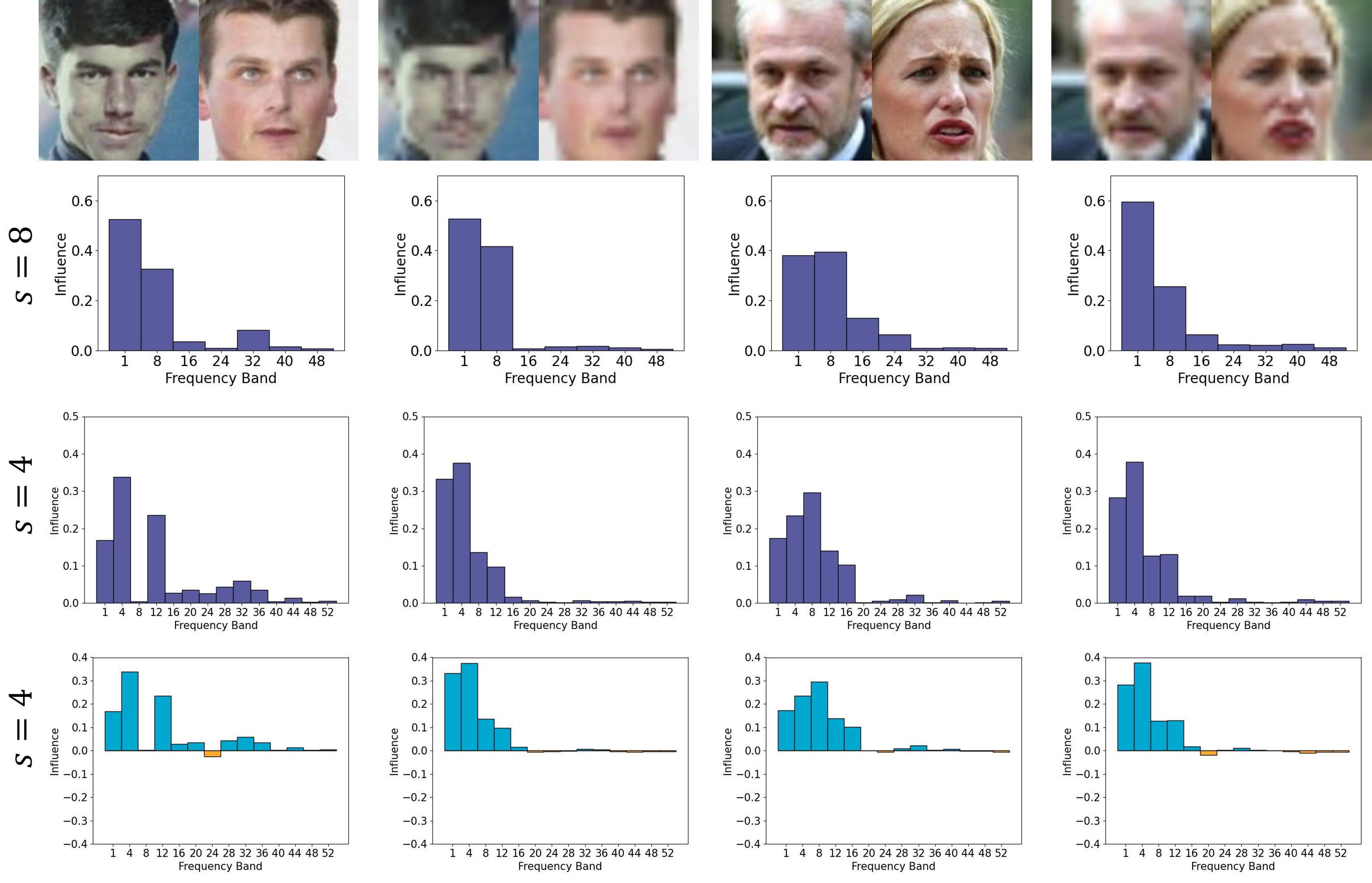}
    \caption{Non-matching pairs and their low-resolution versions with FHPs. Also for the non-matching image pairs the low-frequency bands are most influential for the dissimilarity as captured by the FHPs. In the low-resolution images the low-frequency bands gain even more influence. Under each image pair we show two) with different $s$ absolute FHPs followed by a third directed FHP.}
  \label{fig:qual2}

\end{figure*}

\begin{figure*}[h]
  \centering
  \includegraphics[width=0.8\textwidth]{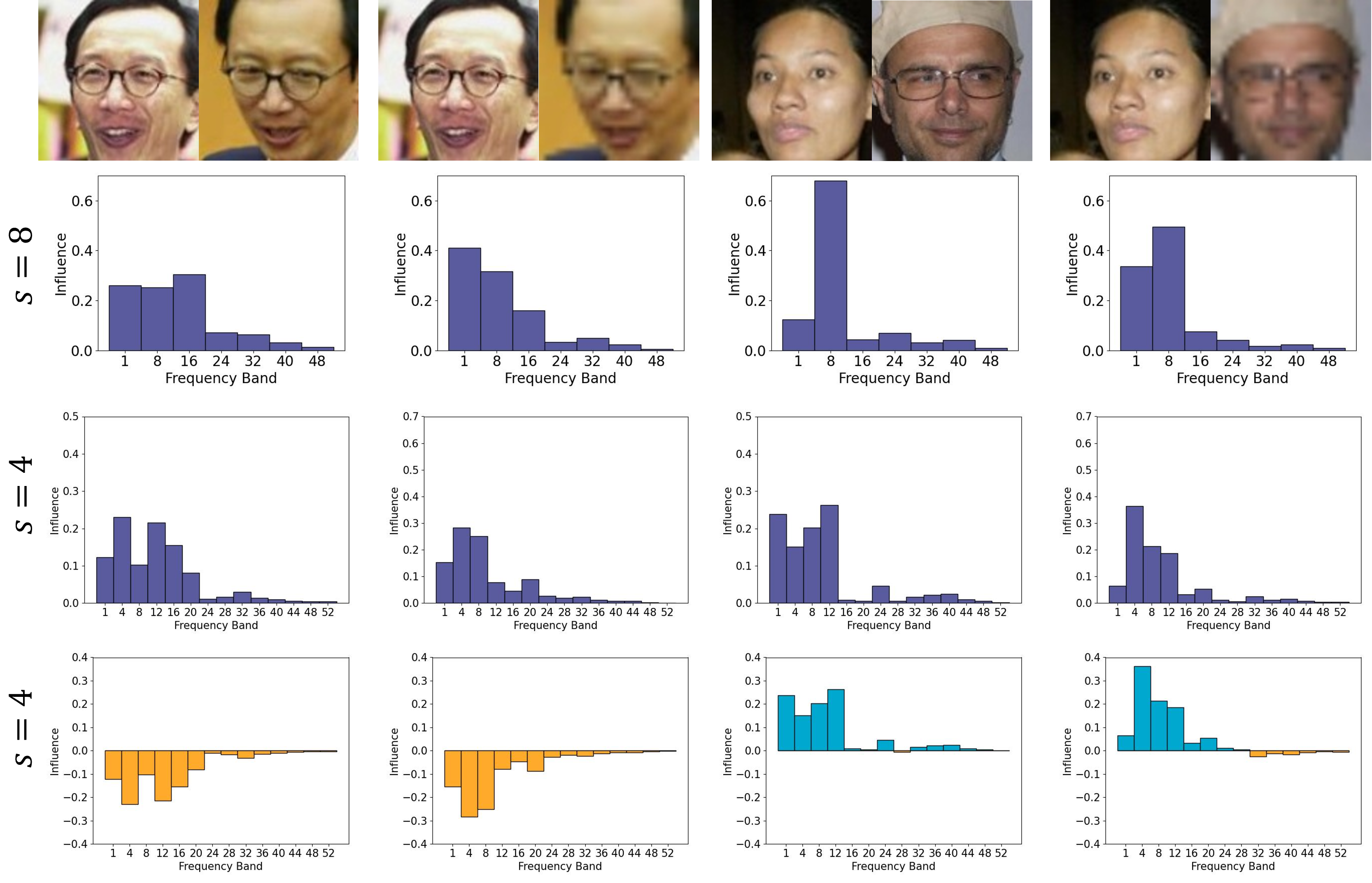}
    \caption{Matching and non-matching pairs with cross-resolution counterpart. In the cross-resolution scenario, also a shift to the lower frequency-bands can be observed, indicating that the FR model relies more on low-frequency information than high-frequency information in this case. Under each image pair we show two) with different $s$ absolute FHPs followed by a third directed FHP.}
  \label{fig:qual3}
\end{figure*}

For the visual analysis, we investigate two image pairs from the LFW dataset \cite{lfw}, one genuine (matching) pair and one imposter (non-matching) pair. The investigation is extended to the evaluation of how FHPs change in a low-resolution and cross-resolution case. All the FHPs in this section have been created using the ElasticFace-Arc model \cite{elasticface} and $L2$-norm. We provide more examples in the supplementary material.

Fig. \ref{fig:qual1} shows two matching (genuine) pairs with their corresponding FHPs. The first row of FHPs provides the absolute influence FHP  with frequency band size $s=8$, while the second row shows the absolute influence FHP with frequency band size $s=4$, The last row of FHPs shows the directed FHP with the same band size as the second row. Next to the unaltered pairs (columns 1 and 3), we provide the low-resolution pairs (columns 2 and 4) that have been downscaled and upscaled with a scaling factor of $0.25$. The FHPs in Fig. \ref{fig:qual1} show, that for both unaltered pairs the low to mid frequencies (1 to 28) have the highest influence on the FR model decision, explaining the impact of certain frequency bands on the decision. In contrast, high frequencies (40 to 56) are not as influential. As we expect if our explainability tool functions appropriately, there is a strong agreement between FHPs with different frequency band sizes $s$. The directed FHP in the last row shows that the influence of the frequency bands is negative, meaning that removing the frequency band leads to a decrease in the similarity score. This follows the expectation as removing frequency bands means that frequencies containing identity information are removed which reduces the similarity of the matching pair. Based on the FHPs provided, it also shows that the explanations are image pair-specific. When comparing the FHP of the low-resolution image pair with the FHP of the unaltered image pair, the FHPs of the low-resolution pairs indicate the higher influence of the lower frequencies as it is the case on the FHPs of the unaltered image pairs. This shows that the approach captures that image details are mainly present in higher frequencies that have been altered by the down- and up-sampling of the images.

Two non-matching (imposter) pairs, their low-resolution counterparts, and the corresponding FHPs are shown in Fig. \ref{fig:qual2}. Similar to the matching pairs in Fig. \ref{fig:qual1}, the most influential frequency bands are the low-frequency bands. In contrast to the matching pairs, the similarity increases in most cases for the non-matching pairs as frequency bands are removed (as shown by the directed FHPs). As the pairs are from different identities and the similarity scores should be small (in comparison to genuine pairs), removing certain frequencies can actually lead to a minor increase in similarity. It is important to note, as described in Sec. \ref{sec:vis}, that the FHPs are normalized for each pair to make them comparable and therefore the change in the similarity score might be smaller in the imposter pairs than in the genuine pairs. Similar to the observations in Fig. \ref{fig:qual1}, the FHPs of the low-resolution pairs show a higher influence of the lower frequencies on the similarity score than their counterparts, indicating the lack of discriminative identity details in the low-resolution images. 

In Fig. \ref{fig:qual3}, samples of the cross-resolution scenario are compared to the unaltered image pairs. We provide a matching and non-matching pair and in both cases, the influence shifts even more to the lower frequency bands when investigating the cross-resolution in comparison to the unaltered image pair indicating that the FR model relies on low-frequency information regarding the similarity of the faces.

\section{Conclusion}
In this work, we proposed a novel explainability space for FR by going beyond the explanations in the spatial domain to explanations in the frequency domain. To achieve that, we utilize a frequency-based masking scheme allowing us to assign an influence score to each frequency component (band). Recent work has shown, that CNNs which are the backbone of current FR systems, utilize different frequencies including frequencies imperceptible to humans to come to their conclusion. Our approach moves away from spatial heatmaps to explain the behavior of these models in the frequency domain using frequency heat plots (FHPs), obtaining more complete explanations. Our quantitative and visual evaluation using two state-of-the-art FR models showed the effectiveness of our proposed approach also investigating two use cases. Furthermore, being the first work to propose frequency-based explanations of verification-based decisions, we aim to motivate more novel approaches in this direction, even beyond FR. Possible future work could include the combination of spatial and frequency-based explanations.

{\small
\bibliographystyle{ieee_fullname}
\bibliography{1_paper}
}

\end{document}


\title{Beyond Spatial Explanations: Explainable Face Recognition in the Frequency Domain - Supplementary Material}

\author{Marco Huber$^{1,2}$ and Naser Damer$^{1,2}$\\
$^{1}$ Fraunhofer Institute for Computer Graphics Research IGD, Darmstadt, Germany\\
$^{2}$ Department of Computer Science, TU Darmstadt,
Darmstadt, Germany\\
Email: marco.huber@igd.fraunhofer.de
}
\maketitle

\section{Introduction}
This is the supplementary material to the paper: \textit{Beyond Spatial Explanations: Explainable Face Recognition in the Frequency Domain}. In this supplementary material, we provide additional quantitative results, the approach applied to the use case of morphing attacks, and more visual examples and investigations. While we provided the insertion and deletion curves using the $\mathcal{L}_{2}$-norm following Wang \etal [34] as the norm in Equation (2) in the paper, we provide here the insertion and deletion curves using the $\mathcal{L}_{1}$-norm as proposed by Abello \etal [1]. For the visual investigation and examples, we provide additional samples using different bandsizes, the two different face recognition models (ElasticFace-Arc [3], CurricularFace [10]), and the two different norms ($\mathcal{L}_{1}$, $\mathcal{L}_{2}$).

\section{Qualitative Results using $L_{2}$-norm}
\begin{figure}
     \centering
     \begin{subfigure}[b]{0.35\textwidth}
         \centering
         \includegraphics[width=\textwidth]{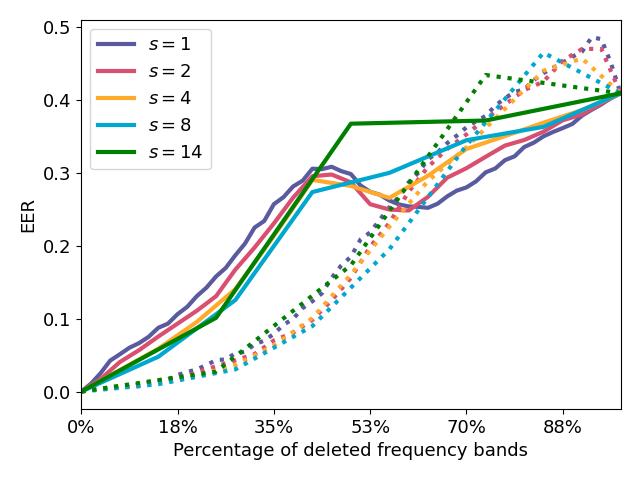}
         \caption{Deletion - EER}
         \label{fig:elaDelEER}
     \end{subfigure}
     \begin{subfigure}[b]{0.35\textwidth}
         \centering
         \includegraphics[width=\textwidth]{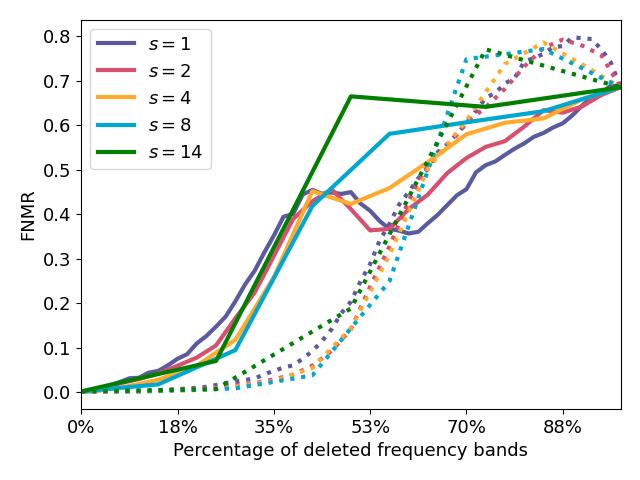}
         \caption{Deletion - FNMR}
         \label{fig:elaDelFNMR}
     \end{subfigure}
     \begin{subfigure}[b]{0.35\textwidth}
         \centering
         \includegraphics[width=\textwidth]{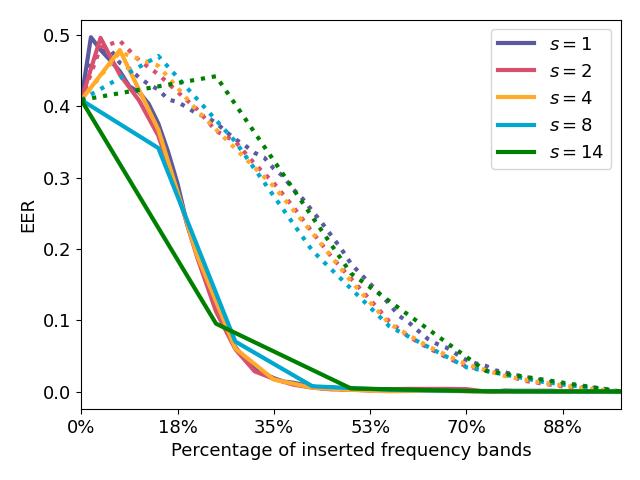}
         \caption{Insertion - EER}
         \label{fig:ealInsEER}
     \end{subfigure}
     \begin{subfigure}[b]{0.35\textwidth}
         \centering
         \includegraphics[width=\textwidth]{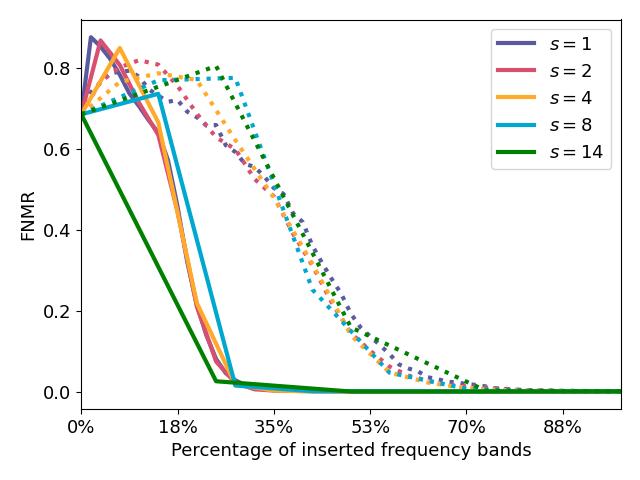}
         \caption{Insertion - FNMR}
         \label{fig:elaInsFNMR}
     \end{subfigure}
        \caption{Deletion and insertion curves using ElasticFace-Arc [3] and $\mathcal{L}_{1}$-norm while Figure 3 in the main paper uses $\mathcal{L}_{2}$-norm. The solid lines are a result of our proposed explanations. The dotted line indicates the performance of the baseline with the same frequency band size $s$ as its color counterpart in solid line. Both faster ascending deletion curves and faster descending insertion curves point the effectiveness of the proposed explanations. }
        \label{fig:quantElastic}
\end{figure}

\begin{figure}
     \centering
     \begin{subfigure}[b]{0.35\textwidth}
         \centering
         \includegraphics[width=\textwidth]{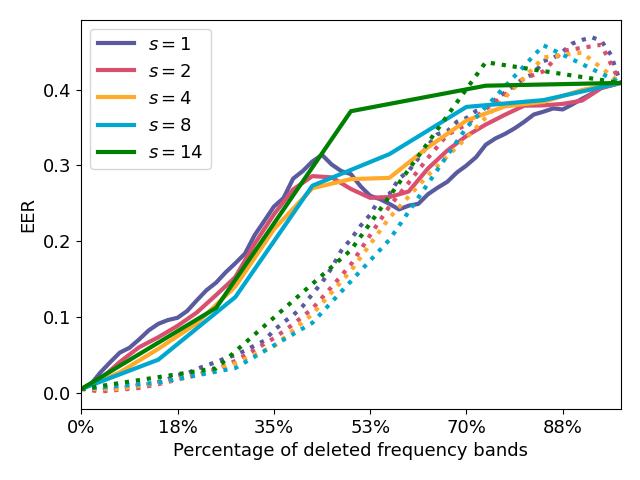}
         \caption{Deletion - EER}
         \label{fig:curDelEER}
     \end{subfigure}
     \begin{subfigure}[b]{0.35\textwidth}
         \centering
         \includegraphics[width=\textwidth]{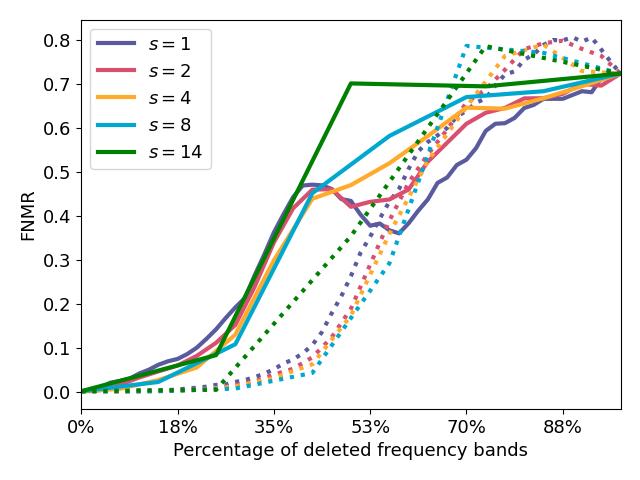}
         \caption{Deletion - FNMR}
         \label{fig:curDelFNMR}
     \end{subfigure}
     \begin{subfigure}[b]{0.35\textwidth}
         \centering
         \includegraphics[width=\textwidth]{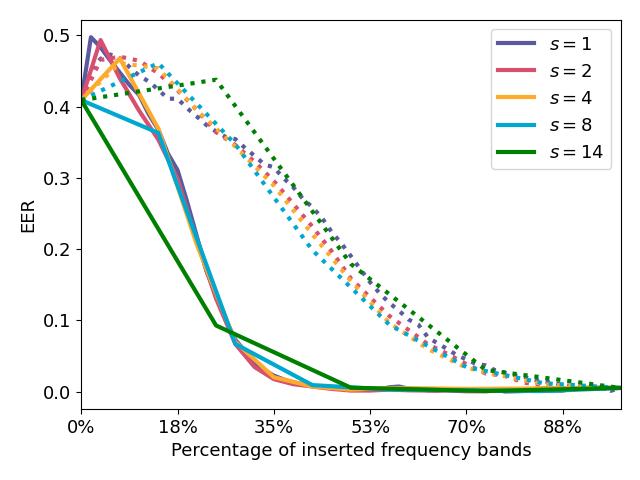}
         \caption{Insertion - EER}
         \label{fig:curInsEER}
     \end{subfigure}
     \begin{subfigure}[b]{0.35\textwidth}
         \centering
         \includegraphics[width=\textwidth]{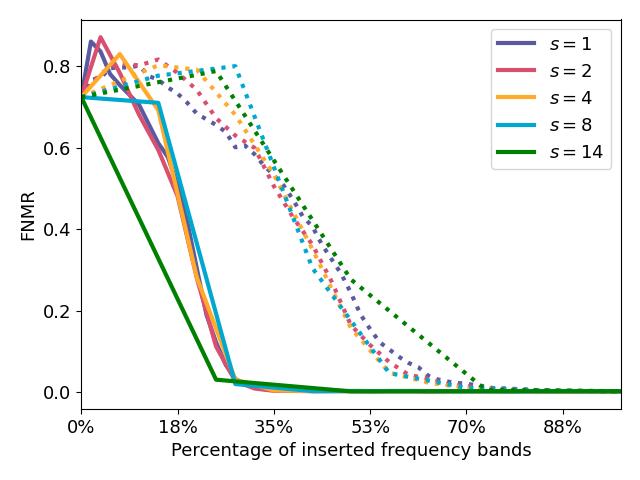}
         \caption{Insertion - FNMR}
         \label{fig:curInsFNMR}
     \end{subfigure}
        \caption{Deletion and insertion curves using CurricularFace [34] and $\mathcal{L}_{1}$-norm while Figure 4 in the main paper uses $\mathcal{L}_{2}$-norm.. The solid lines are a result of our proposed explanations. The dotted line indicates the performance of the baseline with the same frequency band size $s$ as its color counterpart in solid line. Both faster ascending deletion curves and faster descending insertion curves point the effectiveness of the proposed explanations.}
        \label{fig:quantCurr}
\end{figure}

Due to space we only provided the qualitative results masking utilizing $\mathcal{L}_{2}$-norm as proposed by Wang \etal [34]. In contrast, Abello \etal [1] proposed to use $\mathcal{L}_{1}$-norm rather than $\mathcal{L}_{2}$-norm for masking in the frequency domain due to the better suitability for discrete spaces [1]. The results are provided in Fig. \ref{fig:quantElastic} for ElasticFace-Arc [3] and in Fig. \ref{fig:quantCurr} for CurricularFace [10]. Similar to the results presented in the paper using $\mathcal{L}_{2}$-norm, the faster increasing deletion curves and the faster increasing insertion curves prove the effectiveness of our approach also when using $\mathcal{L}_{1}$-norm with only slightly difference between the different norms.


\subsection{Use Case: Morphing Attacks}
\begin{figure*}[htbp]
    \centering
    \begin{subfigure}[b]{0.24\textwidth}
        \centering
        \includegraphics[width=\textwidth]{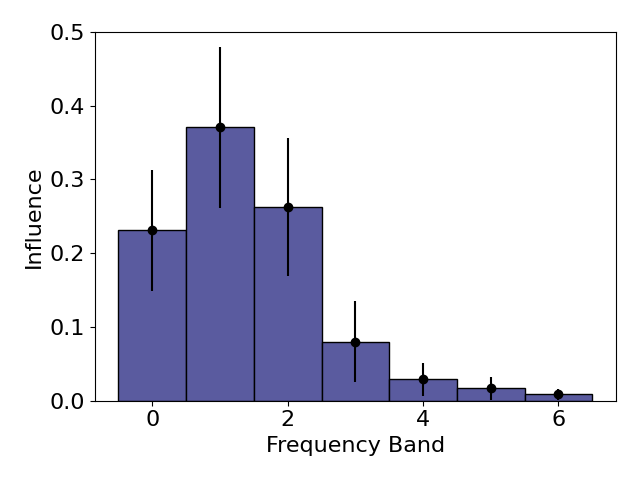}
        \caption{Bona fide}
        \label{fig:bf}
    \end{subfigure}
    \begin{subfigure}[b]{0.24\textwidth}
        \centering
        \includegraphics[width=\textwidth]{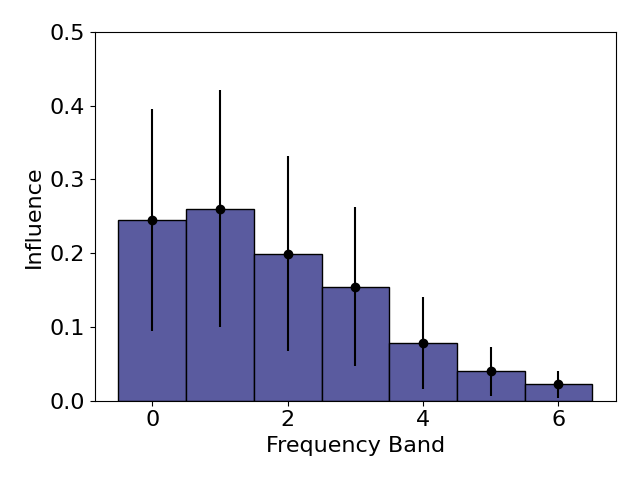}
        \caption{MorDIFF}
        \label{fig:Diff}
    \end{subfigure}
        \begin{subfigure}[b]{0.24\textwidth}
        \centering
        \includegraphics[width=\textwidth]{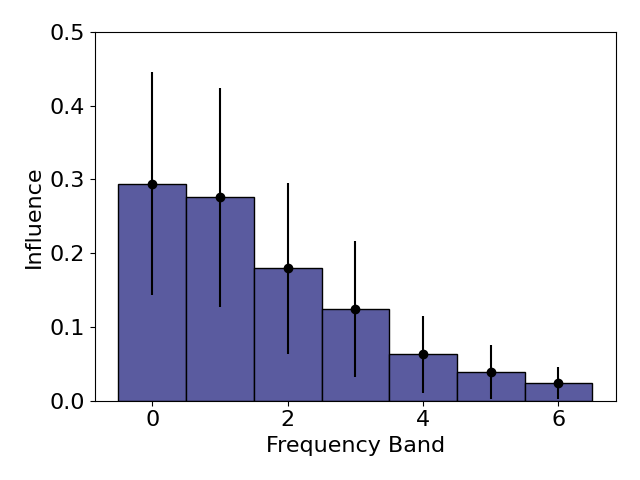}
        \caption{MIPGAN2}
        \label{fig:mip}
    \end{subfigure}
    \begin{subfigure}[b]{0.24\textwidth}
        \centering
        \includegraphics[width=\textwidth]{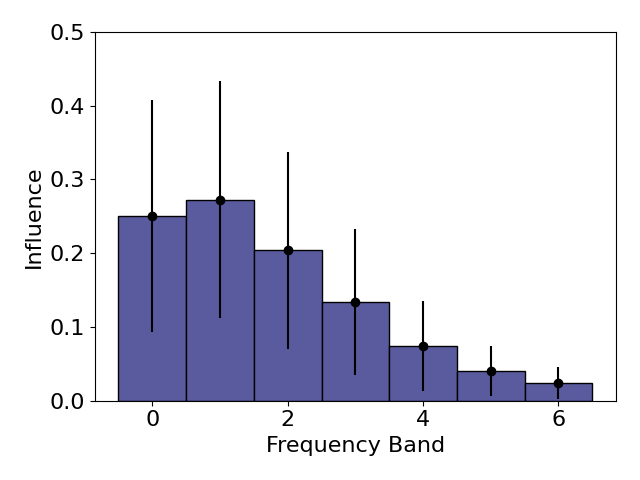}
        \caption{OpenCV}
        \label{fig:OpenCV}
    \end{subfigure}
    \caption{Mean absolute FHPs for bona fide pairs and three different morphing attacks. In morphing attack pairs, the frequency importance is more distributed compared to the bona fide importance and the variety is higher (error bar, standard deviation).}
    \label{fig:morphing}
\end{figure*}

As a second use case in addition to the provided cross-resolution use case presented in the main paper, we investigate explaining the frequency importance of bona fide and morphing attack comparison to references with the assumption that the morphing process would effect some frequency component of the image, and thus its comparison to the reference. For the morphing attack images, we utilize the SYN-MAD 2022 dataset \cite{DBLP:conf/icb/HuberBLRRDNGSCT22} which consists of around 1000 morphing images of each morphing approach and 204 bona fide images. In Figure \ref{fig:morphing}, we provide the mean absolute FHPs plot and the standard deviation for the FHPs with $s = 8$. The figures shows that the frequency importance is wider distributed for all three morphing attack types (OpenCV (landmark-based), MIPGAN2 (GAN-based) \cite{DBLP:journals/tbbis/ZhangVRRDB21}, MorDIFF (diffusion-based) \cite{DBLP:conf/iwbf/DamerFSKHB23}) in comparison to the bona fide FHPs and also show higher variety (shown as the error bar, standard deviation). This allows us to conclude based on the proposed frequency-based explanations, that FR models process morphing attack differently than unaltered bona fide images. These findings support the usability of the frequency-based explanations in diverse use cases to gain new insights into FR model behavior.

\section{Visual Investigations }
\subsection{Comparison of  $L_{1}$- and $L_{2}$-norm}
\begin{figure*}
    \centering
    \includegraphics[width=\textwidth]{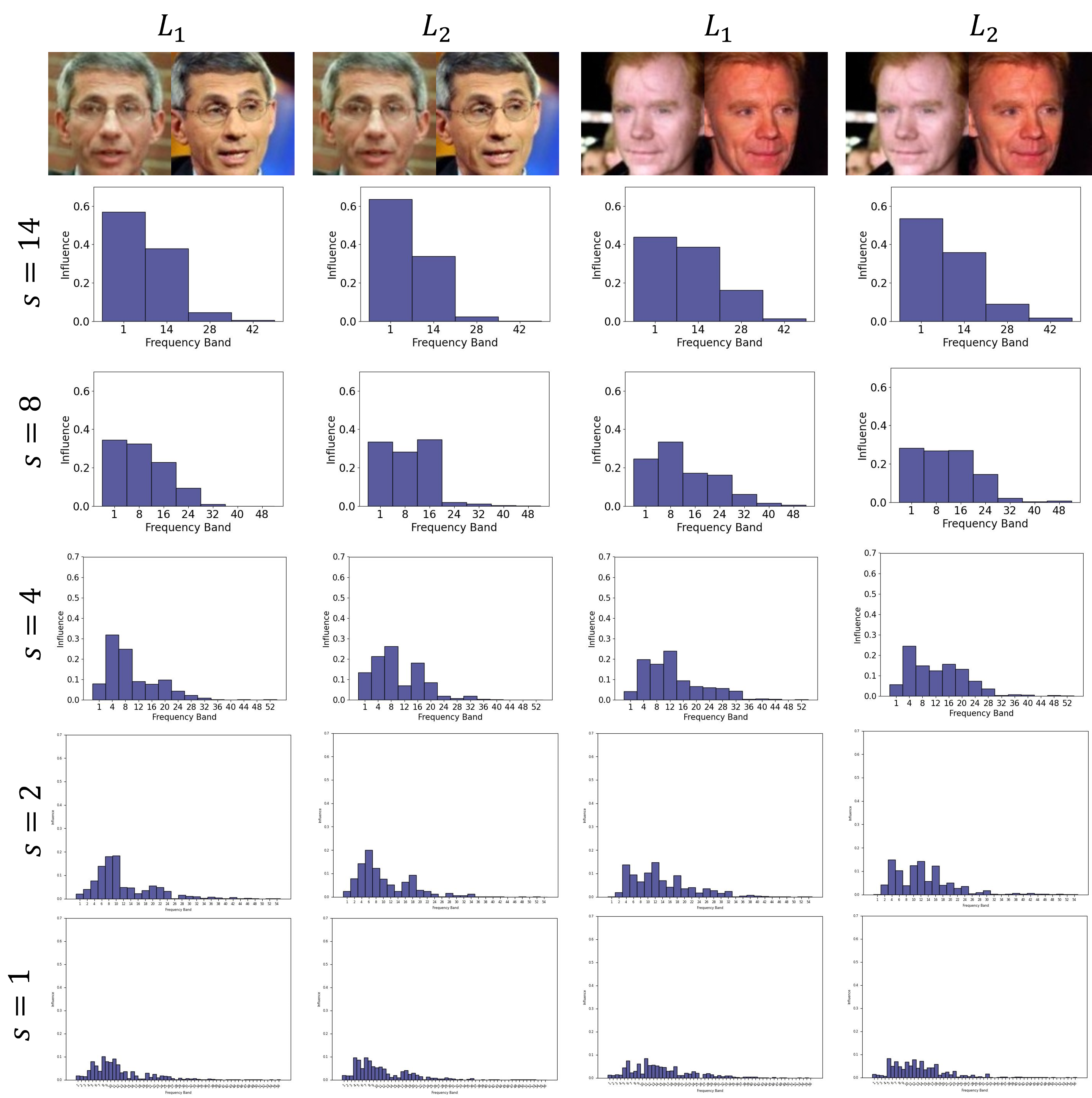}
    \caption{Absolute FHPs using the ElasticFace-Arc [3] model on unaltered matching pairs and either $\mathcal{L}_{1}$- or $\mathcal{L}_{2}$-norm during the masking process. The distribution of the influences show some similarity independent of the norm utilized for masking, especially with smaller bandsizes. For the sake of comparison, we keep the y-axis scale fixed over all absolute FHPs.}
   \label{fig:matchElasticL1L2LFWAbs}
\end{figure*}

\begin{figure*}
    \centering
    \includegraphics[width=\textwidth]{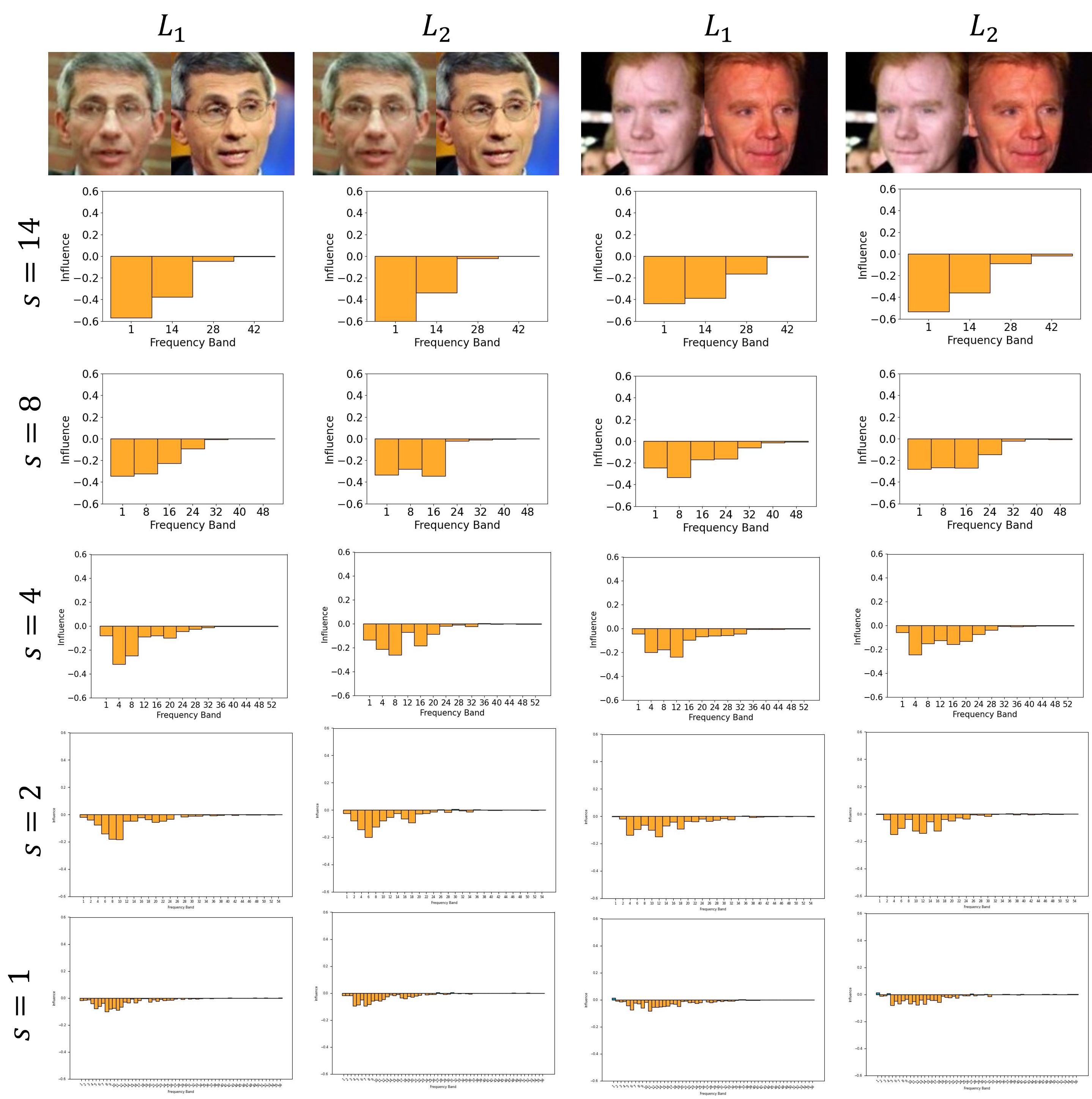}
    \caption{Directed FHPs using the ElasticFace-Arc [3] model on unaltered matching pairs and either $\mathcal{L}_{1}$- or $\mathcal{L}_{2}$-norm during the masking process. The distribution of the influences show some similarity independent of the norm utilized for masking, especially with smaller bandsizes. For the sake of comparison, we keep the y-axis scale fixed over all directed FHPs.}
   \label{fig:matchElasticL1L2LFWDir}
\end{figure*}

\begin{figure*}
    \centering
    \includegraphics[width=\textwidth]{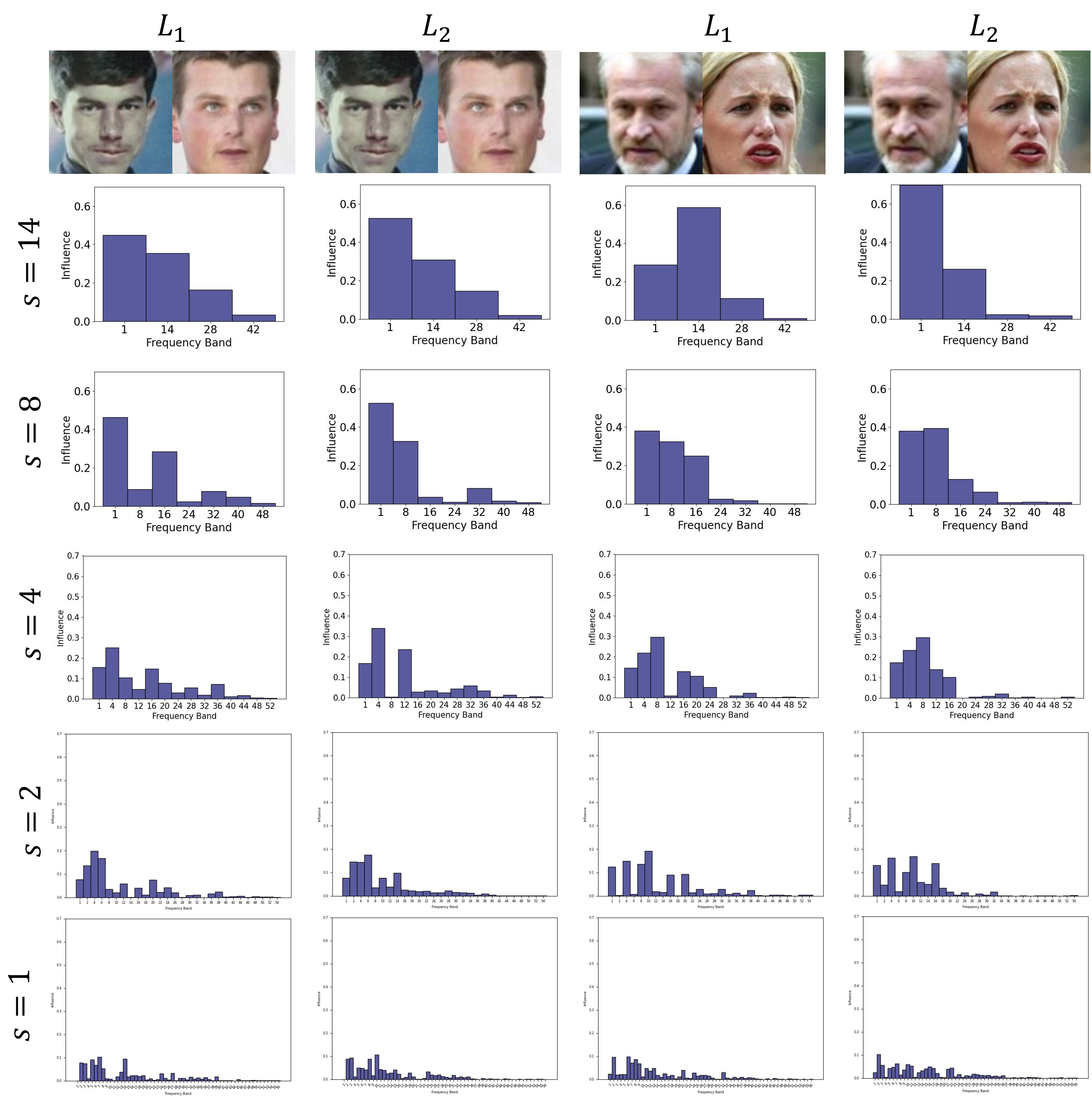}
    \caption{Absolute FHPs using the ElasticFace-Arc [3] model on low-resolution non-matching pairs and either $\mathcal{L}_{1}$- or $\mathcal{L}_{2}$-norm during the masking process. The distribution of the influences show some similarity independent of the norm utilized for masking, especially with smaller bandsizes. For the sake of comparison, we keep the y-axis scale fixed over all absolute FHPs.}
   \label{fig:nomatchElaL1L2LFWAbs}
\end{figure*}

\begin{figure*}
    \centering
    \includegraphics[width=\textwidth]{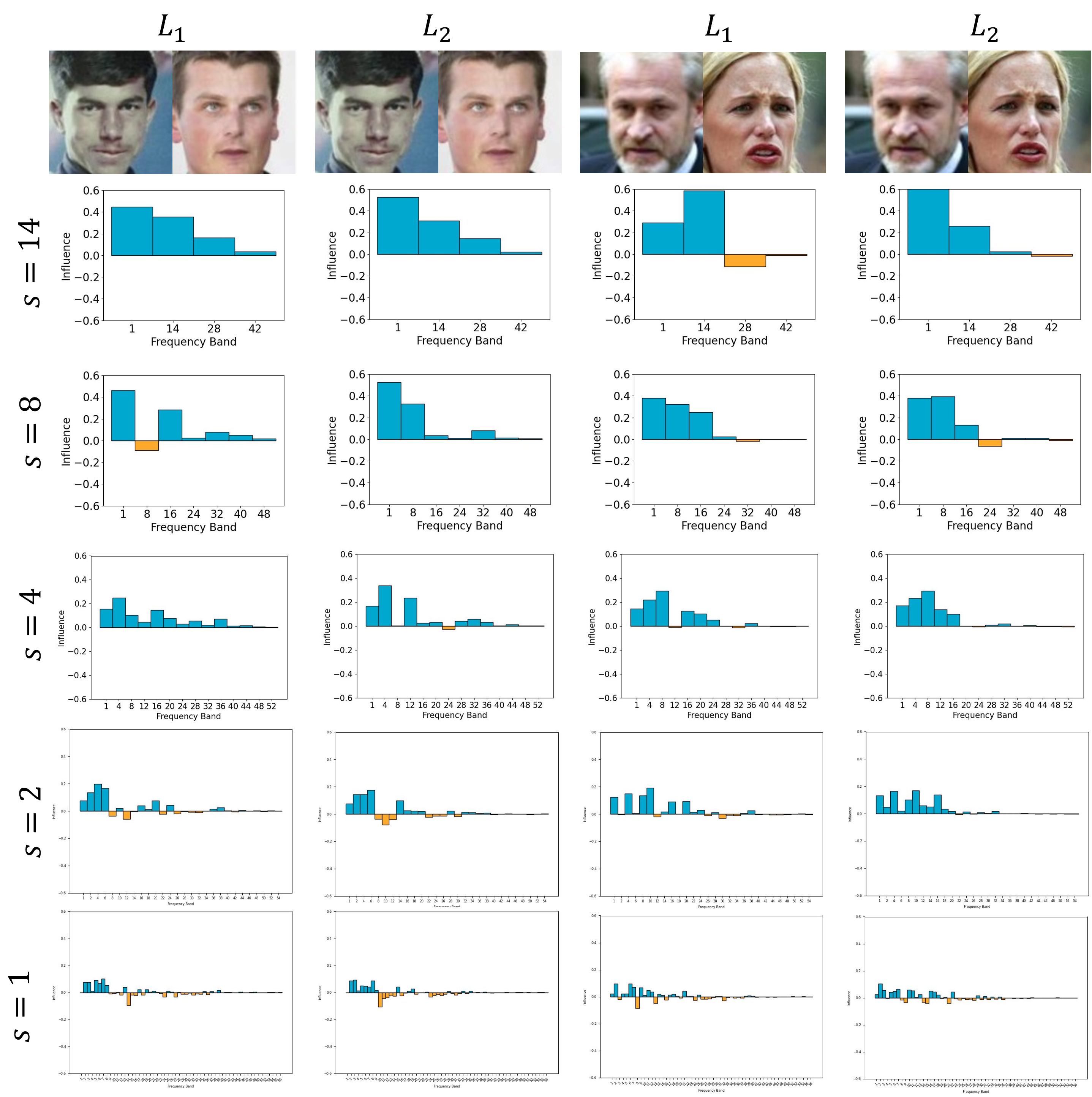}
    \caption{Directed FHPs using the ElasticFace-Arc [3] model on low-resolution non-matching pairs and either $\mathcal{L}_{1}$- or $\mathcal{L}_{2}$-norm during the masking process. The distribution of the influences show some similarity independent of the norm utilized for masking, especially with smaller bandsizes. For the sake of comparison, we keep the y-axis scale fixed over all directed FHPs.}
   \label{fig:nomatchElaL1L2LFWDir}
\end{figure*}

\begin{figure*}
    \centering
    \includegraphics[width=\textwidth]{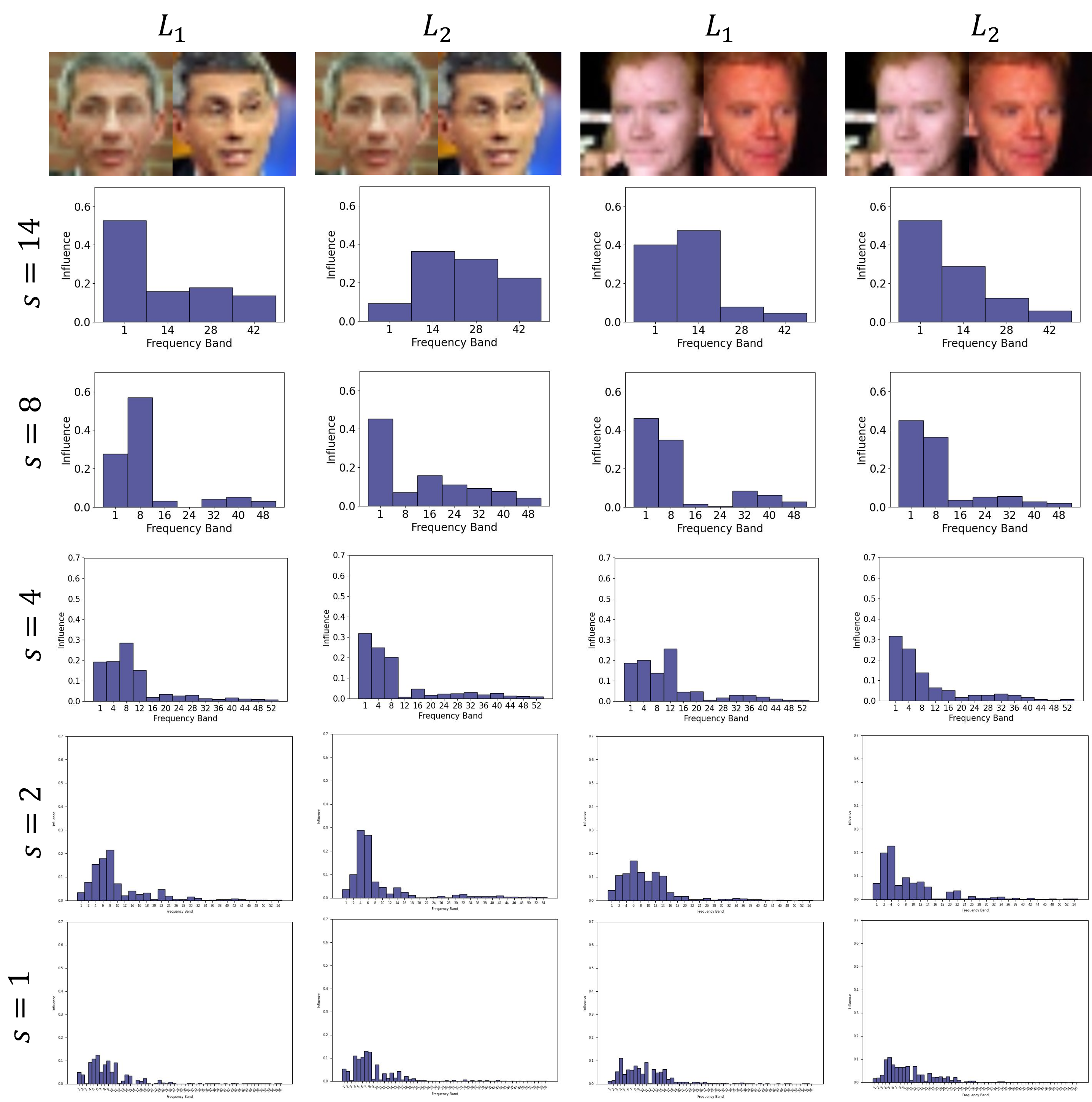}
    \caption{Absolute FHPs using the ElasticFace-Arc [3] model on low-resolution matching pairs and either $\mathcal{L}_{1}$- or $\mathcal{L}_{2}$-norm during the masking process. The distribution of the influences show some similarity independent of the norm utilized for masking, especially with smaller bandsizes. For the sake of comparison, we keep the y-axis scale fixed over all absolute FHPs.}
   \label{fig:matchElasticL1L2LowAbs}
\end{figure*}

\begin{figure*}
    \centering
    \includegraphics[width=\textwidth]{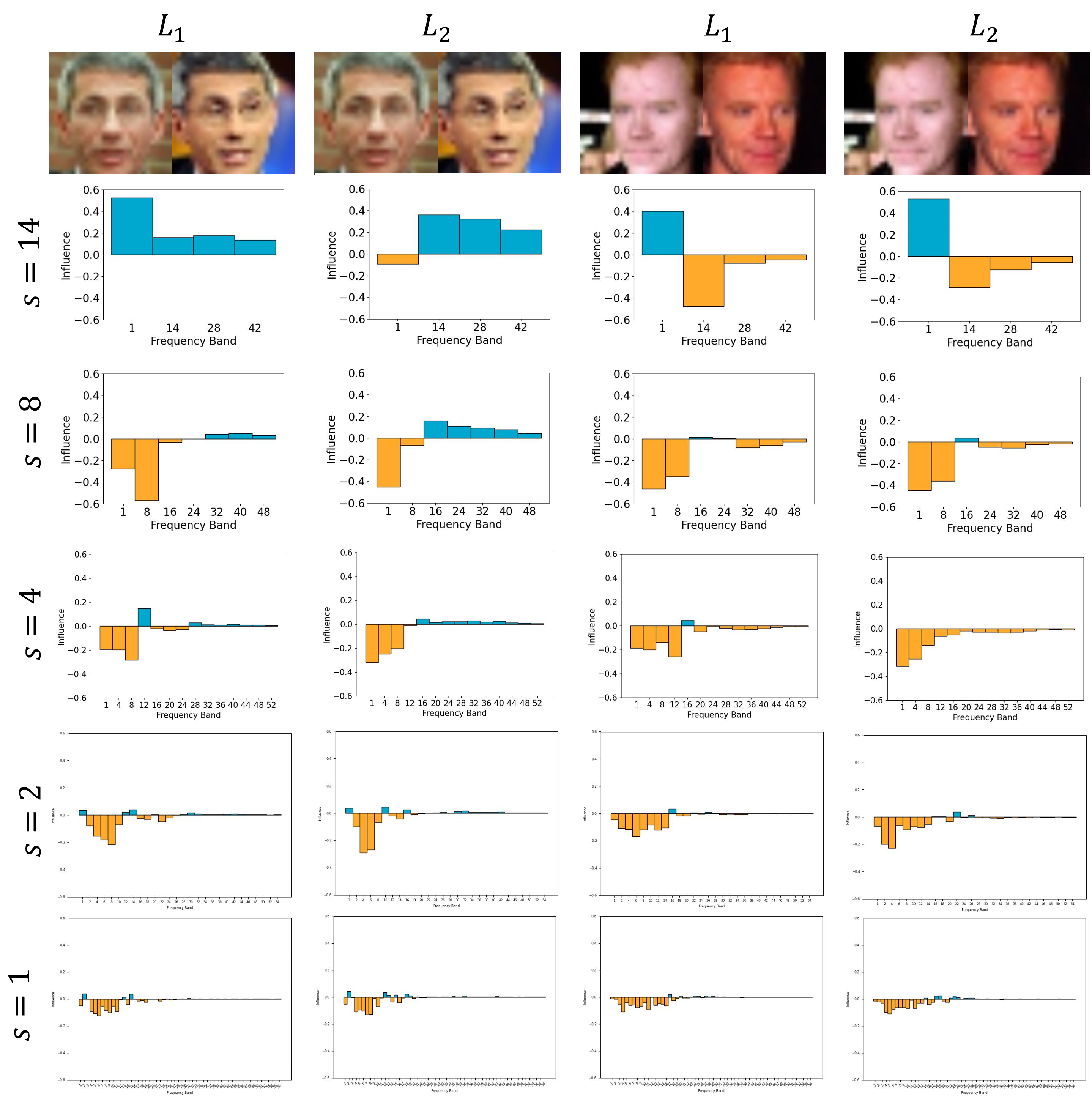}
    \caption{Directed FHPs using the ElasticFace-Arc [3] model on low-resolution matching pairs and either $\mathcal{L}_{1}$- or $\mathcal{L}_{2}$-norm during the masking process. The distribution of the influences show some similarity independent of the norm utilized for masking, especially with smaller bandsizes. For the sake of comparison, we keep the y-axis scale fixed over all directed FHPs.}
   \label{fig:matchElasticL1L2LowDir}
\end{figure*}

\begin{figure*}
    \centering
    \includegraphics[width=\textwidth]{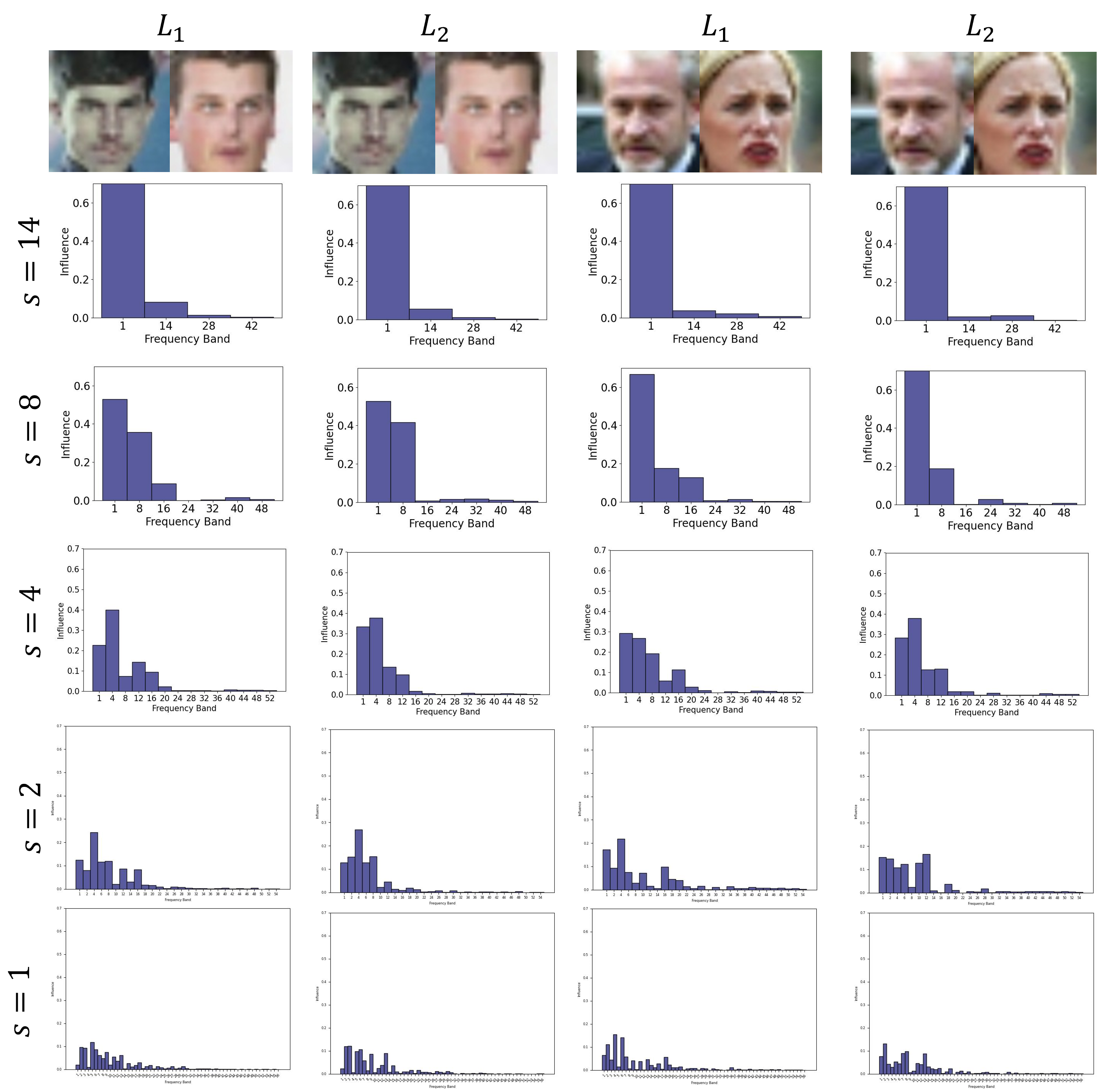}
    \caption{Absolute FHPs using the ElasticFace-Arc [3] model on low-resolution non-matching pairs and either $\mathcal{L}_{1}$- or $\mathcal{L}_{2}$-norm during the masking process. The distribution of the influences show some similarity independent of the norm utilized for masking, especially with smaller bandsizes. For the sake of comparison, we keep the y-axis scale fixed over all absolute FHPs.}
   \label{fig:nomatchElasticL1L2LowAbs}
\end{figure*}

\begin{figure*}
    \centering
    \includegraphics[width=\textwidth]{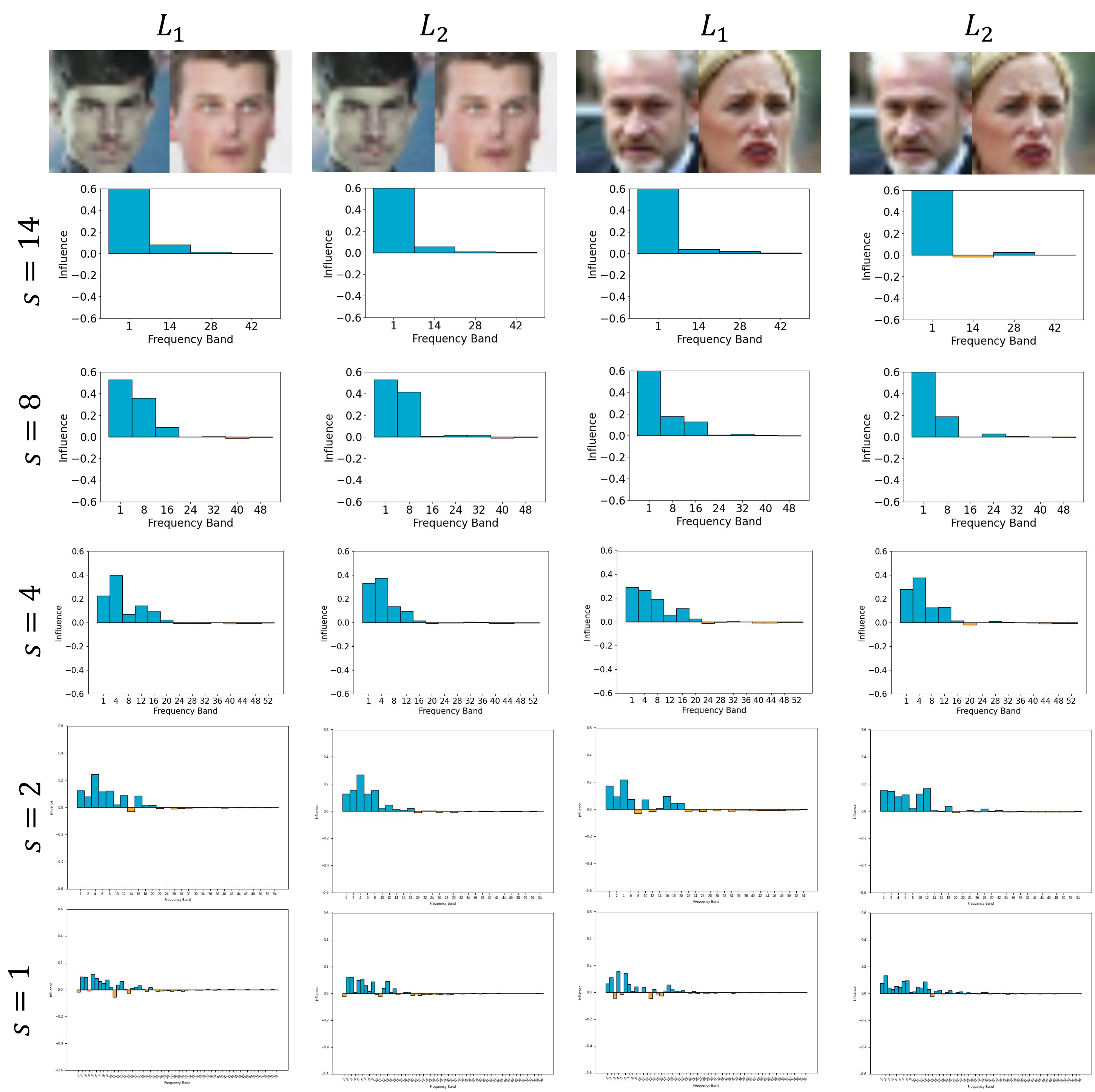}
    \caption{Directed FHPs using the ElasticFace-Arc [3] model on low-resolution non-matching pairs and either $\mathcal{L}_{1}$- or $\mathcal{L}_{2}$-norm during the masking process. The distribution of the influences show some similarity independent of the norm utilized for masking, especially with smaller bandsizes. For the sake of comparison, we keep the y-axis scale fixed over all directed FHPs.}
   \label{fig:nomatchElaL1L2lowDir}
\end{figure*}

Due to space limitations, we provided in the paper only FHPs using $\mathcal{L}_{2}$-norm during the masking process of our approach. Here we present the absolute and directed Frequency Heat Plots (FHPs) utilizing either $\mathcal{L}_{1}$- or $\mathcal{L}_{2}$-norm on the unaltered and the low-resolution image pairs for both, matches and non-matches. We also include the FHPs for all investigated bandsizes $s \in [1,2,4,8,14]$. The results for the unaltered pairs are presented in Figures \ref{fig:matchElasticL1L2LFWAbs}, \ref{fig:matchElasticL1L2LFWDir}, \ref{fig:nomatchElaL1L2LFWAbs}, and \ref{fig:nomatchElaL1L2LFWDir}. The results on low-resolution image pairs are presented in Figures \ref{fig:matchElasticL1L2LowAbs}, \ref{fig:matchElasticL1L2LowDir}, \ref{fig:nomatchElasticL1L2LowAbs} and \ref{fig:nomatchElaL1L2lowDir}. Similar to the observation in the quantitative analysis above, the assigned influences for the different frequency bands is quite similar, independent of the utilized norm.

\subsection{Comparison between ElasticFace-Arc and CurricularFace}
\begin{figure*}
    \centering
    \includegraphics[width=\textwidth]{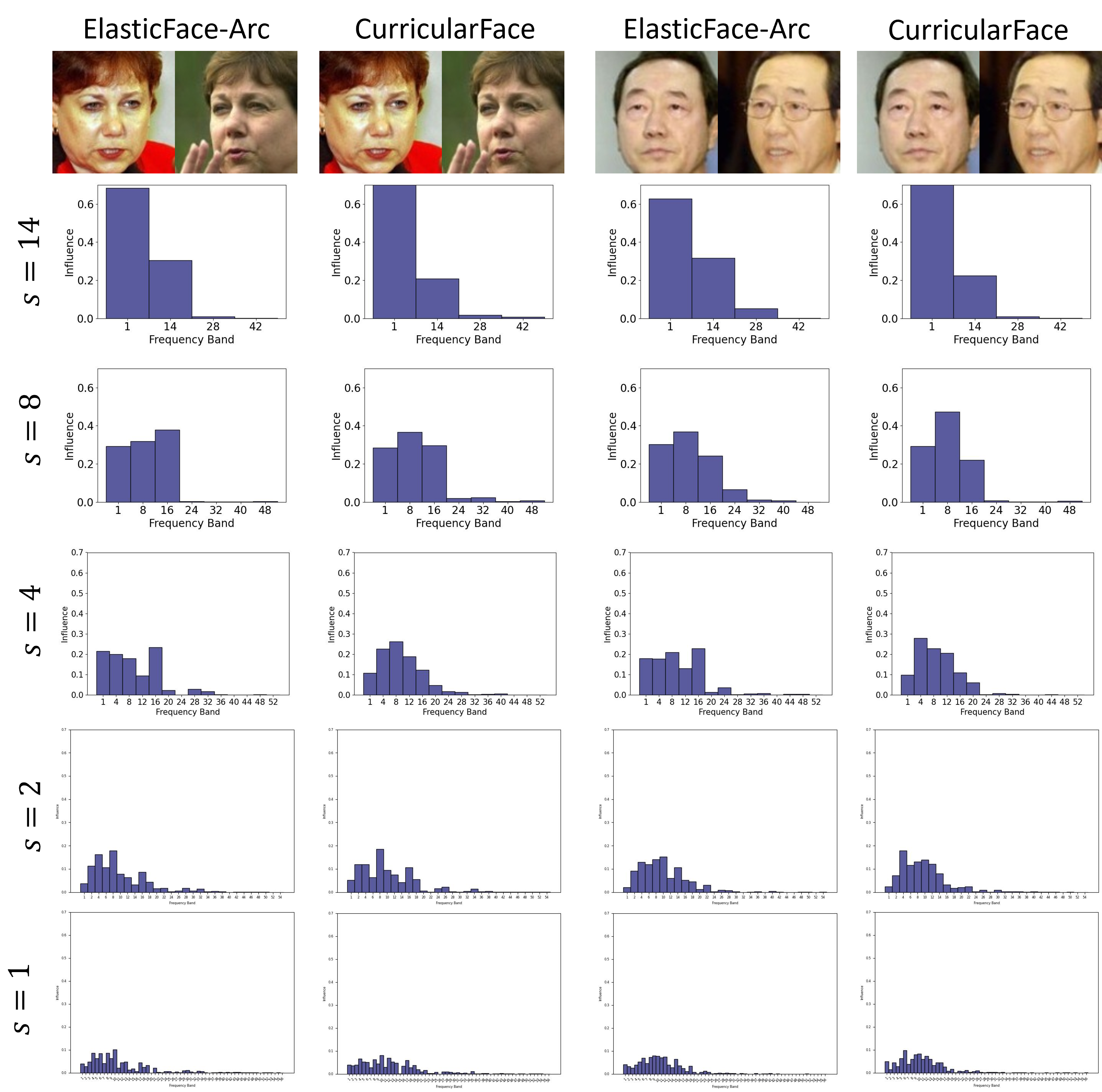}
    \caption{Comparison of absolute FHPs using the ElasticFace-Arc [3] model and the CurricularFace [10] model on unaltered matching pairs using $\mathcal{L}_{2}$-norm during the masking process. The distribution of the influences show that, to some extend, the same frequency bands have a similar influence independent from the utilized model. For the sake of comparison, we keep the y-axis scale fixed over all absolute FHPs.}
   \label{fig:matchModelLFWabs}
\end{figure*}

\begin{figure*}
    \centering
    \includegraphics[width=\textwidth]{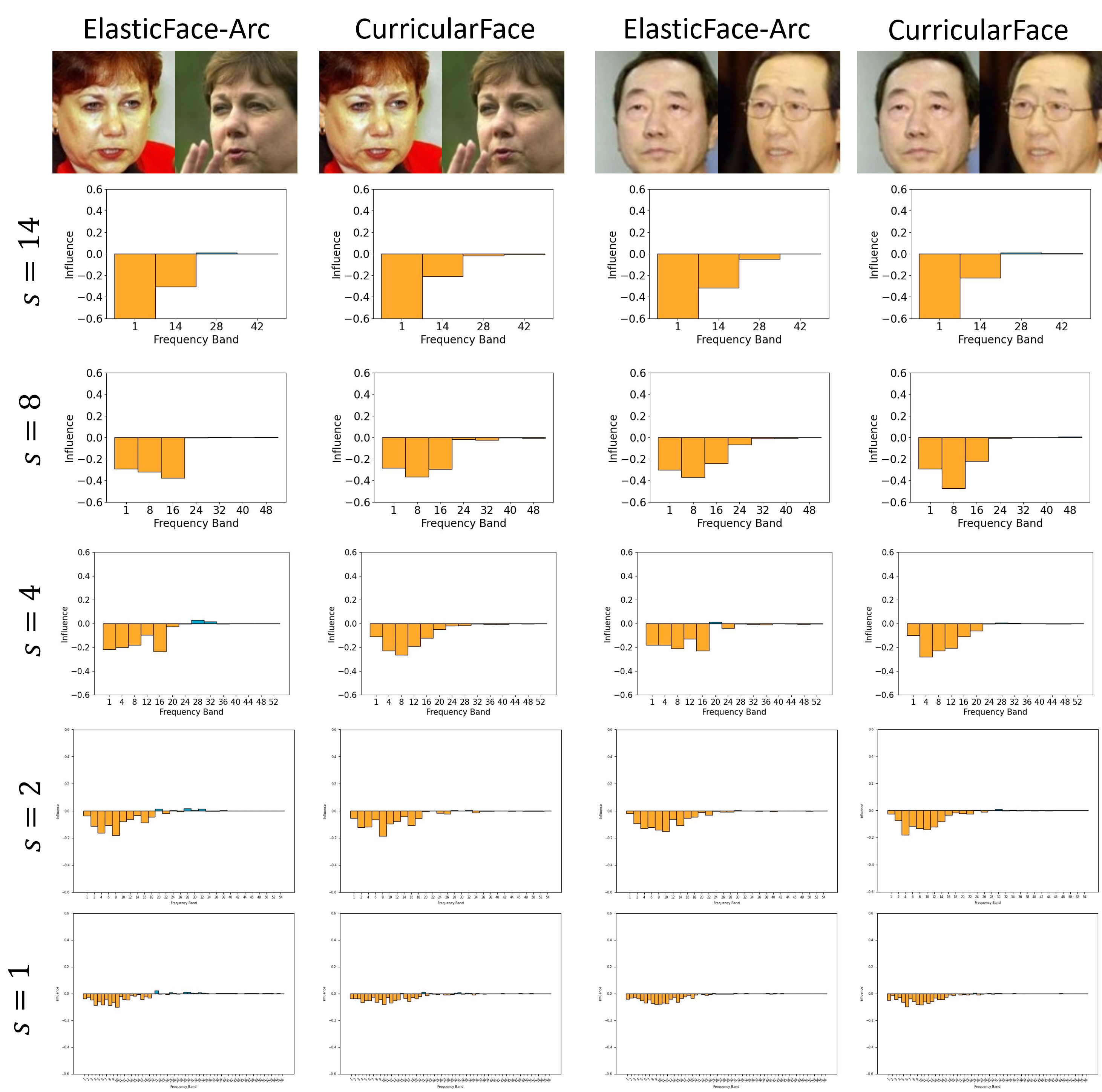}
    \caption{Comparison of directed FHPs using the ElasticFace-Arc [3] model and the CurricularFace [10] model on unaltered matching pairs using $\mathcal{L}_{2}$-norm during the masking process. The distribution of the influences show that, to some extend, the same frequency bands have a similar influence independent from the utilized model. For the sake of comparison, we keep the y-axis scale fixed over all directed FHPs.}
   \label{fig:matchModelLFWdir}
\end{figure*}

\begin{figure*}
    \centering
    \includegraphics[width=\textwidth]{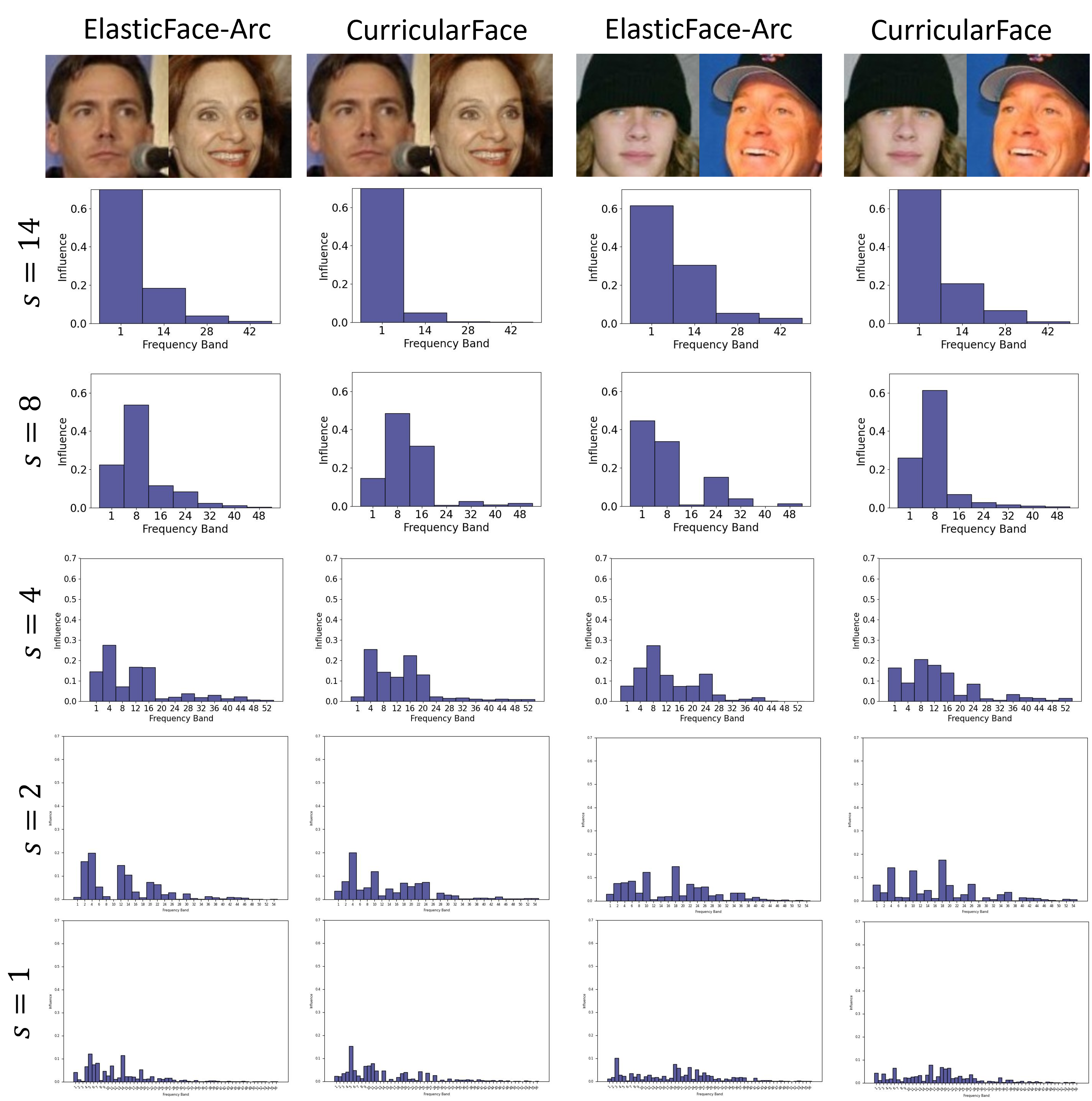}
    \caption{Comparison of absolute FHPs using the ElasticFace-Arc [3] model and the CurricularFace [10] model on unaltered non-matching pairs using $\mathcal{L}_{2}$-norm during the masking process. The distribution of the influences show that, to some extend, the same frequency bands have a similar influence independent from the utilized model. For the sake of comparison, we keep the y-axis scale fixed over all absolute FHPs.}
   \label{fig:nomatchModelLFWabs}
\end{figure*}

\begin{figure*}
    \centering
    \includegraphics[width=\textwidth]{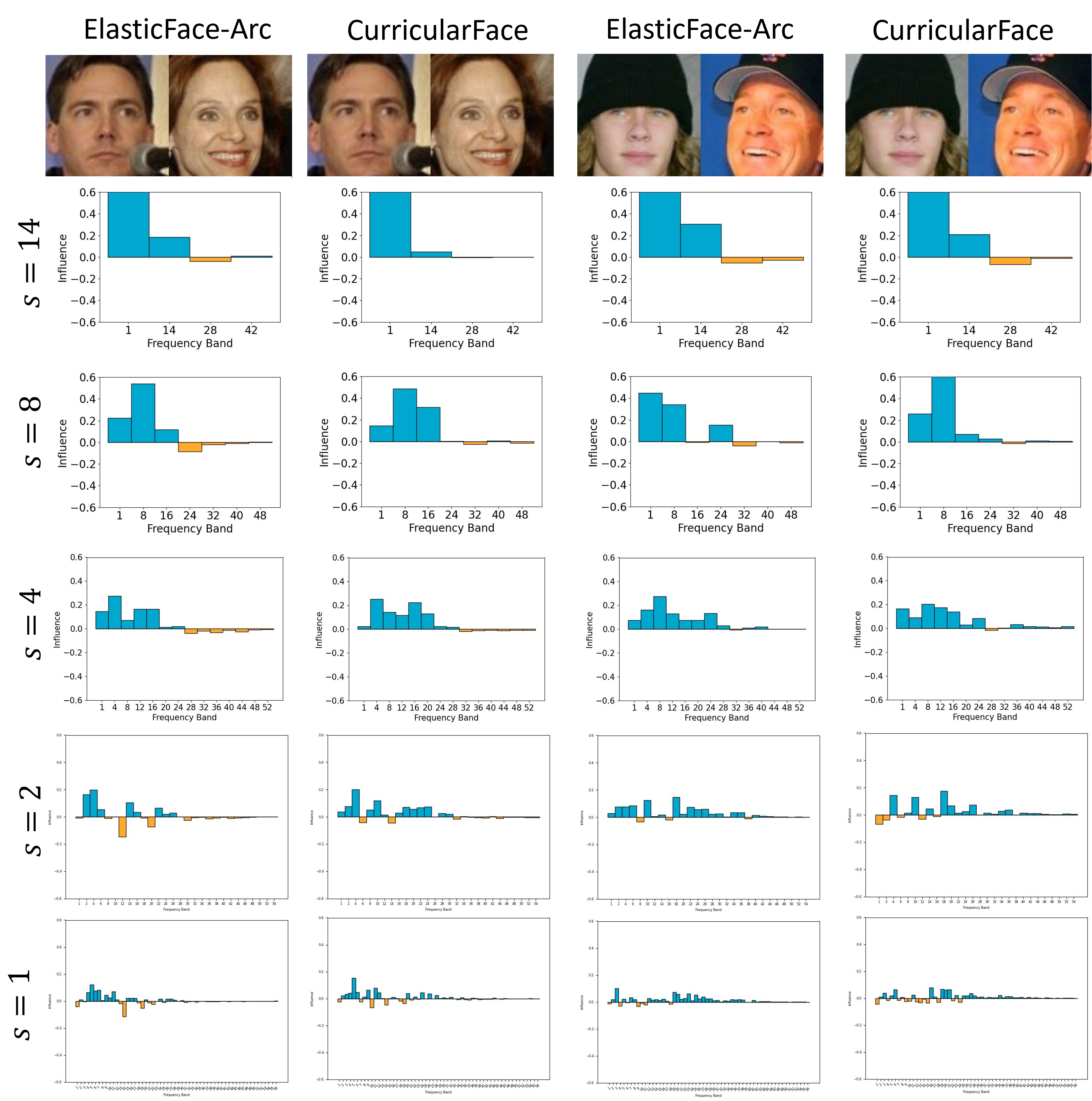}
    \caption{Comparison of directed FHPs using the ElasticFace-Arc [3] model and the CurricularFace [10] model on unaltered non-matching pairs using $\mathcal{L}_{2}$-norm during the masking process. The distribution of the influences show that, to some extend, the same frequency bands have a similar influence independent from the utilized model. For the sake of comparison, we keep the y-axis scale fixed over all directed FHPs.}
   \label{fig:nomatchModelLFWdir}
\end{figure*}

\begin{figure*}
    \centering
    \includegraphics[width=\textwidth]{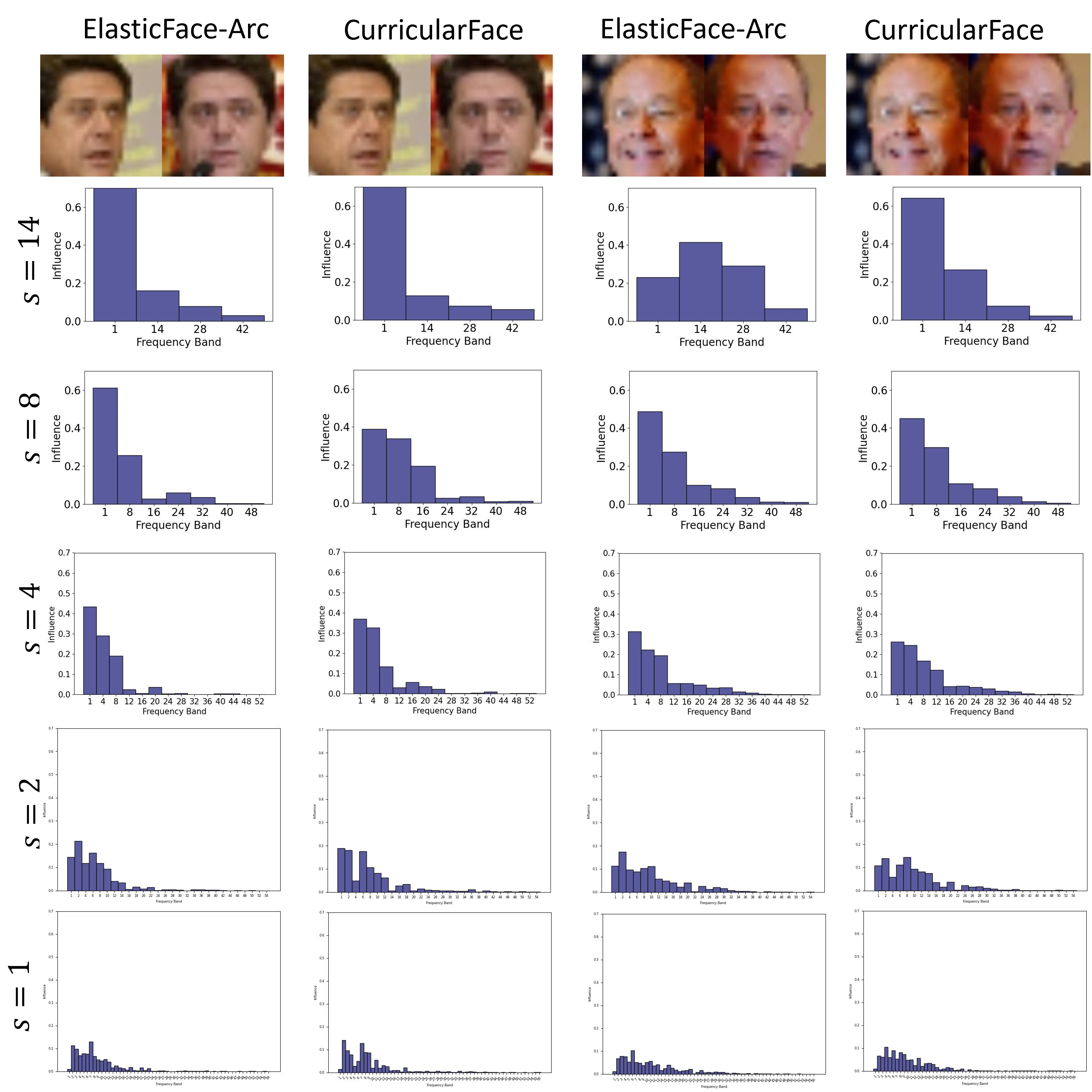}
    \caption{Comparison of absolute FHPs using the ElasticFace-Arc [3] model and the CurricularFace [10] model on low-resolution matching pairs using $\mathcal{L}_{2}$-norm during the masking process. The distribution of the influences show that, to some extend, the same frequency bands have a similar influence independent from the utilized model. For the sake of comparison, we keep the y-axis scale fixed over all absolute FHPs.}
   \label{fig:matchModelLowAbs}
\end{figure*}

\begin{figure*}
    \centering
    \includegraphics[width=\textwidth]{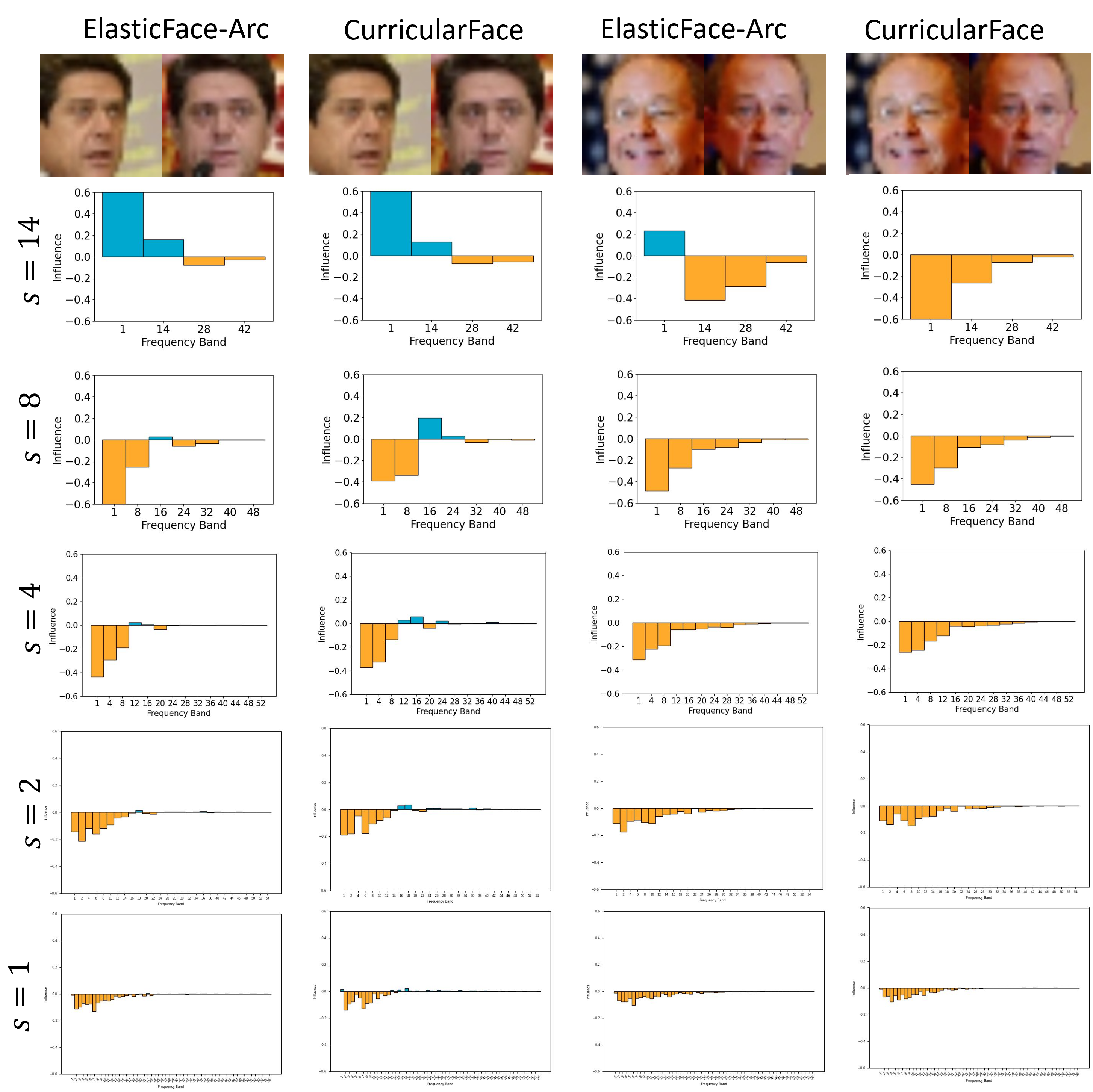}
    \caption{Comparison of directed FHPs using the ElasticFace-Arc [3] model and the CurricularFace [10] model on low-resolution matching pairs using $\mathcal{L}_{2}$-norm during the masking process. The distribution of the influences show that, to some extend, the same frequency bands have a similar influence independent from the utilized model. For the sake of comparison, we keep the y-axis scale fixed over all directed FHPs.}
   \label{fig:matchModelLowdir}
\end{figure*}

\begin{figure*}
    \centering
    \includegraphics[width=\textwidth]{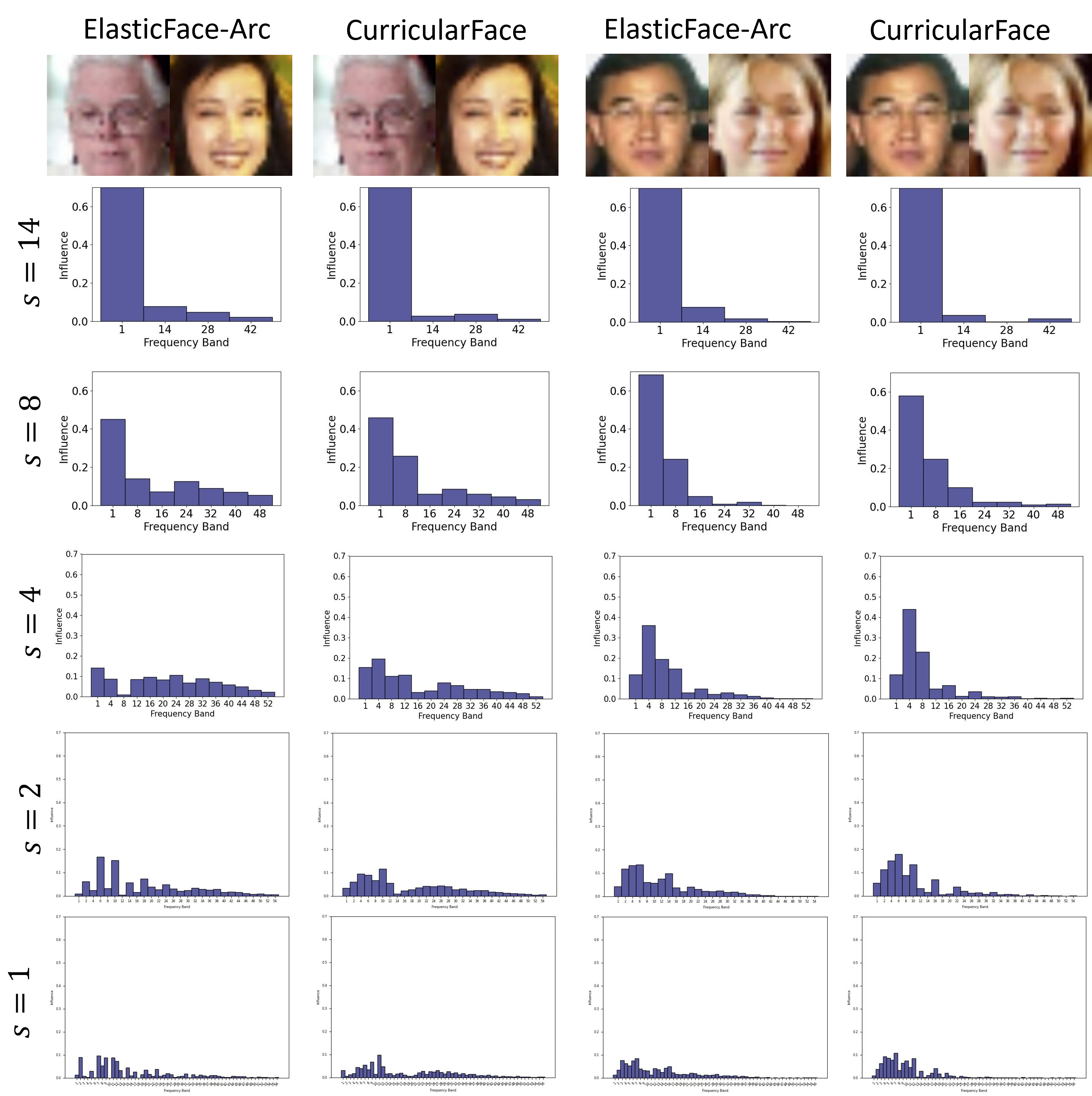}
    \caption{Comparison of absolute FHPs using the ElasticFace-Arc [3] model and the CurricularFace [10] model on low-resolution non-matching pairs using $\mathcal{L}_{2}$-norm during the masking process. The distribution of the influences show that, to some extend, the same frequency bands have a similar influence independent from the utilized model. For the sake of comparison, we keep the y-axis scale fixed over all absolute FHPs.}
   \label{fig:nomatchModelLowAbs}
\end{figure*}

\begin{figure*}
    \centering
    \includegraphics[width=\textwidth]{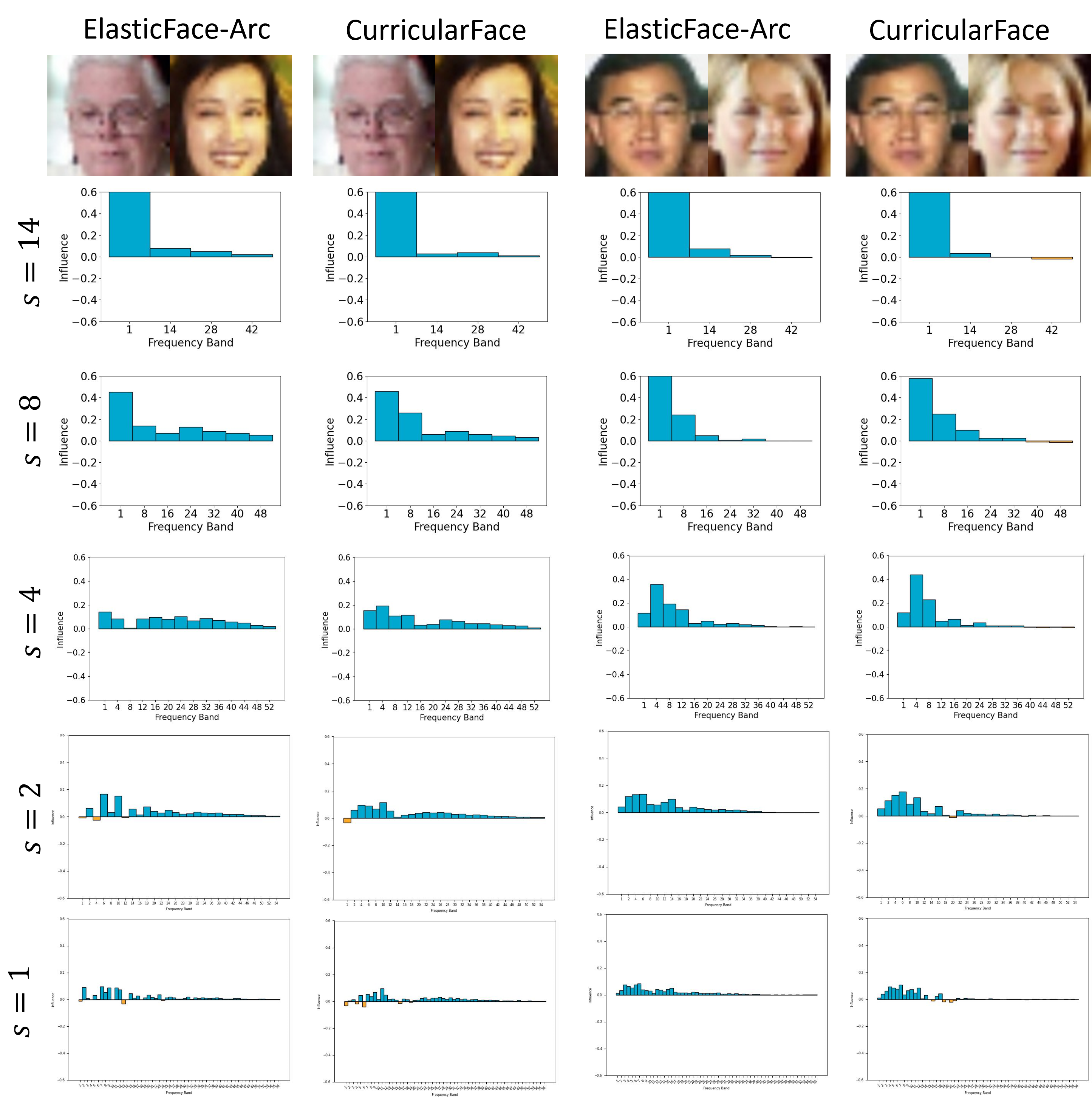}
    \caption{Comparison of directed FHPs using the ElasticFace-Arc [3] model and the CurricularFace [10] model on low-resolution non-matching pairs using $\mathcal{L}_{2}$-norm during the masking process. The distribution of the influences show that, to some extend, the same frequency bands have a similar influence independent from the utilized model. For the sake of comparison, we keep the y-axis scale fixed over all directed FHPs.}
   \label{fig:nomatchModelLowdir}
\end{figure*}

We provide in the paper the quantitative results for both FR models and some FHPs utilizing the ElasticFace-Arc [3] model (due to limited space). Here, we additionally provide some examples as an comparison between the absolute and directed FHPs on matching and non-matching image pairs, as well as unaltered and low-resolution image pairs. The FHPs based on both models for matching pairs are provided in the Figures \ref{fig:matchModelLFWabs}, \ref{fig:matchModelLFWdir}, \ref{fig:nomatchModelLFWabs} and \ref{fig:nomatchModelLFWdir}. For the non-matching pairs, some FHPs based on both models are provided in the Figures \ref{fig:matchModelLowAbs}, \ref{fig:matchModelLowdir}, \ref{fig:nomatchModelLowAbs} and \ref{fig:nomatchModelLowdir}. The FHPs provided show, that to some extend, the same frequency bands are similarly influential on the image pairs, independent from the model utilized.

\let\oldthebibliography=\thebibliography
\let\oldendthebibliography=\endthebibliography
\renewenvironment{thebibliography}[1]{%
    \oldthebibliography{#1}%
    \setcounter{enumiv}{ 40 }%
}{\oldendthebibliography}
{\small
\bibliographystyle{ieee_fullname}
\bibliography{2_supp}
}